\PassOptionsToPackage{unicode}{hyperref}
\PassOptionsToPackage{hyphens}{url}
\documentclass[
  11pt,
]{article}
\usepackage{xcolor}
\usepackage[margin=1in]{geometry}
\usepackage{amsmath,amssymb}
\usepackage{iftex}
\ifPDFTeX
  \usepackage[T1]{fontenc}
  \usepackage[utf8]{inputenc}
  \usepackage{textcomp} 
\else 
  \usepackage{unicode-math} 
  \defaultfontfeatures{Scale=MatchLowercase}
  \defaultfontfeatures[\rmfamily]{Ligatures=TeX,Scale=1}
\fi
\usepackage{lmodern}
\ifPDFTeX\else
\fi
\IfFileExists{upquote.sty}{\usepackage{upquote}}{}
\IfFileExists{microtype.sty}{
  \usepackage[]{microtype}
  \UseMicrotypeSet[protrusion]{basicmath} 
}{}
\makeatletter
\@ifundefined{KOMAClassName}{
  \IfFileExists{parskip.sty}{%
    \usepackage{parskip}
  }{
    \setlength{\parindent}{0pt}
    \setlength{\parskip}{6pt plus 2pt minus 1pt}}
}{
  \KOMAoptions{parskip=half}}
\makeatother
\usepackage{longtable,booktabs,array}
\usepackage{calc} 
\usepackage{etoolbox}
\makeatletter
\patchcmd\longtable{\par}{\if@noskipsec\mbox{}\fi\par}{}{}
\makeatother
\IfFileExists{footnotehyper.sty}{\usepackage{footnotehyper}}{\usepackage{footnote}}
\makesavenoteenv{longtable}
\usepackage{graphicx}
\makeatletter
\newsavebox\pandoc@box
\newcommand*\pandocbounded[1]{
  \sbox\pandoc@box{#1}%
  \Gscale@div\@tempa{\textheight}{\dimexpr\ht\pandoc@box+\dp\pandoc@box\relax}%
  \Gscale@div\@tempb{\linewidth}{\wd\pandoc@box}%
  \ifdim\@tempb\p@<\@tempa\p@\let\@tempa\@tempb\fi
  \ifdim\@tempa\p@<\p@\scalebox{\@tempa}{\usebox\pandoc@box}%
  \else\usebox{\pandoc@box}%
  \fi%
}
\def\fps@figure{htbp}
\makeatother
\usepackage[numbers,super]{natbib}
\usepackage{graphicx}
\usepackage{booktabs}
\usepackage{float}
\usepackage{amsmath}
\usepackage{textcomp}
\usepackage{caption}
\DeclareUnicodeCharacter{00D7}{\ensuremath{\times}}      
\DeclareUnicodeCharacter{2264}{\ensuremath{\leq}}        
\DeclareUnicodeCharacter{2265}{\ensuremath{\geq}}        
\DeclareUnicodeCharacter{2212}{\ensuremath{-}}           
\DeclareUnicodeCharacter{03BA}{\ensuremath{\kappa}}       
\DeclareUnicodeCharacter{207B}{\textsuperscript{$-$}}    
\DeclareUnicodeCharacter{2070}{\textsuperscript{0}}      
\DeclareUnicodeCharacter{00B9}{\textsuperscript{1}}      
\DeclareUnicodeCharacter{00B2}{\textsuperscript{2}}      
\DeclareUnicodeCharacter{00B3}{\textsuperscript{3}}      
\DeclareUnicodeCharacter{2074}{\textsuperscript{4}}      
\DeclareUnicodeCharacter{2075}{\textsuperscript{5}}      
\DeclareUnicodeCharacter{2076}{\textsuperscript{6}}      
\DeclareUnicodeCharacter{2077}{\textsuperscript{7}}      
\DeclareUnicodeCharacter{2078}{\textsuperscript{8}}      
\DeclareUnicodeCharacter{2079}{\textsuperscript{9}}      

\usepackage{authblk}

\author[1]{Minh-Ha Nguyen}
\author[2]{Erica T. Gray}
\author[2]{Bryce A. Schuler}
\author[3]{Kevin W. Byram}
\author[4]{Chih-Ting Yang}
\author[5]{Fan Ma}
\author[5]{Hua Xu}
\author[6]{Wu-Chen Su}
\author[6]{Chao Yan}
\author[6]{Wei-Qi Wei}
\author[3,6]{Adam Wright}
\author[6]{Lisa Bastarache}
\author[3,6]{Josh F. Peterson}
\author[7]{Lingyao Li}
\author[4]{Siyuan Ma}
\author[ ]{Undiagnosed Diseases Network}
\author[2]{Rizwan Hamid}
\author[2,*]{Thomas A. Cassini}
\author[2,4,6,*]{Cathy Shyr}
\affil[1]{Department of Epidemiology, Vanderbilt University, Nashville, TN, USA}
\affil[2]{Department of Pediatrics, Vanderbilt University Medical Center, Nashville, TN, USA}
\affil[3]{Department of Medicine, Vanderbilt University Medical Center, Nashville, TN, USA}
\affil[4]{Department of Biostatistics, Vanderbilt University Medical Center, Nashville, TN, USA}
\affil[5]{Department of Biomedical Informatics and Data Science, Yale School of Medicine, New Haven, CT, USA}
\affil[6]{Department of Biomedical Informatics, Vanderbilt University Medical Center, Nashville, TN, USA}
\affil[7]{School of Information, University of South Florida, Tampa, FL, USA}
\affil[*]{Co-last and co-corresponding authors. Correspondence: Thomas A. Cassini (thomas.a.cassini@vumc.org) and Cathy Shyr (cathy.shyr@vumc.org)}
\renewcommand{\author}[2][]{}
\usepackage{bookmark}
\IfFileExists{xurl.sty}{\usepackage{xurl}}{} 
\makeatletter
\@ifundefined{xmpquote}{}{}
\makeatother
\hypersetup{
  pdftitle={Teaching agentic AI to generalize expert diagnostic reasoning in rare diseases},
  hidelinks,
  pdfcreator={LaTeX via pandoc}}

\title{Teaching agentic AI to generalize expert diagnostic reasoning in
rare diseases}
\author{}
\date{}

\begin{document}
\maketitle

\subsection{Abstract}\label{abstract}

Rare disease diagnosis depends on expert reasoning that is scarce and
difficult to transfer. Large language models rank the correct disease
first in only 35.4\% of benchmark cases and often rely on learned
phenotype-disease associations rather than reusable diagnostic reasoning
strategies. We developed liteOdyssey through Policy Iteration with Human
Feedback, a process in which model failures and expert corrections are
iteratively consolidated into a clinician-gated, natural-language policy
executed by a language model. Across 1,243 public benchmark cases
spanning 722 rare diseases, liteOdyssey ranked the correct disease first
in 59.3\% of cases versus 26.5\% without the policy, with comparable
gains in cases involving diseases excluded from policy development. The
same policy transferred across model families and sizes without
retraining. Adaptation of the policy to the Undiagnosed Diseases Network
(UDN) improved diagnostic accuracy among 515 UDN patients, with gains
confirmed by blinded physician adjudication. These results show that
expert reasoning can be externalized into an inspectable and revisable
natural-language policy that generalizes across rare diseases, transfers
across model backbones, and adapts to a real-world patient cohort.

\textbf{Explore liteOdyssey:} \url{https://liteodyssey.org}

\begin{center}
\begin{minipage}{\linewidth}
\centering
\vspace{-0.35\baselineskip}
\textbf{Graphical highlight}\\[0.3em]
\includegraphics[width=0.33\linewidth]{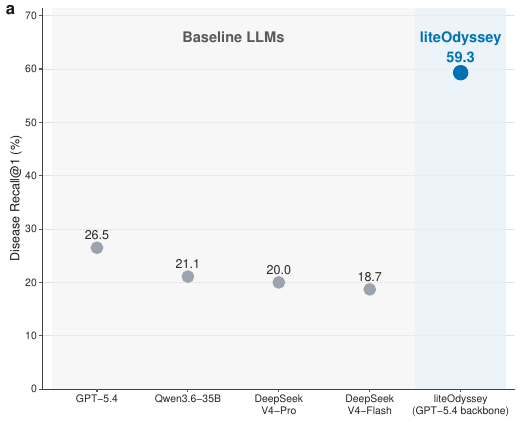}
\end{minipage}
\end{center}

\subsection{Introduction}\label{introduction}

Rare diseases are individually uncommon but collectively affect an
estimated 446 million individuals
worldwide\citep{nguengang_wakap_estimating_2020}. Because individual
rare diseases are unfamiliar to most clinicians, clinically
heterogeneous, and phenotypically overlapping with more common
conditions, affected individuals often experience diagnostic delays
averaging five years, with repeated evaluations, misdiagnoses, and
unnecessary
interventions\citep{marwaha_guide_2022, schieppati_why_2008, faye_time_2024}.
Although recent advances in genetic sequencing have broadened access to
diagnostically relevant evidence, the reasoning process that integrates
clinical and genetic findings to arrive at a diagnosis remains a scarce
expert skill that is difficult to scale, transfer, and apply
consistently across diseases.

Large language models (LLMs) with reasoning capabilities can interpret
phenotypes, weigh evidence, and generate differential diagnoses from
narrative clinical text, yet remain inadequate when used off the shelf:
across public benchmarks, a systematic review and meta-analysis reported
that standalone commercial and specialized biomedical models ranked the
correct disease first in only 35.4\% of cases (95\% CI
30.6--40.4)\citep{nguyen_diagnostic_2026}. Language models are also
susceptible to shortcut learning, relying on surface-level associations
rather than reasoning strategies that transfer to atypical or unfamiliar
cases\citep{geirhos_shortcut_2020}. Current systems improve performance
through domain-specific fine-tuning, solved-case retrieval databases,
knowledge graphs, and multi-agent
orchestration\citep{chen_rarebench_2024, yang_rdguru_2025, yang_specialized_2025, chen_rareagents_2026, zhao_agentic_2026, rose_meddxagent_2025, zheng_end_end_2025, kim_mdagents_2024},
but these approaches can distribute diagnostic reasoning across opaque
model weights, retrieval indices, or orchestration pipelines that are
difficult for clinicians to inspect or revise. Furthermore, these
approaches typically impose additional deployment requirements that may
limit their use in resource-constrained settings.

We hypothesized that current approaches leave underexplored the capacity
of pre-trained language models to execute explicit reasoning procedures
specified in natural language. Pre-trained LLMs can invoke complex task
behavior from natural-language instructions and adapt to new tasks from
few or no examples\citep{radford_language_2019, brown_language_2020}. If
this capacity extends to diagnostic reasoning, expert reasoning could be
externalized as a natural-language policy whose utility extends beyond
the cases and diseases from which it was developed. We therefore asked
whether human-AI collaboration could iteratively construct an explicit
diagnostic policy comprising reusable expert reasoning principles and
actions, and whether that policy could improve rare disease diagnosis
beyond its development set and transfer across different language model
backbones. We further asked whether the approach could be adapted to a
real-world, diagnostically complex patient cohort at the U.S. National
Institutes of Health's Undiagnosed Diseases Network.

We introduce Policy Iteration with Human Feedback (PIHF), a learning
process adapted from generalized policy iteration in reinforcement
learning\citep{sutton_reinforcement_2018}, in which expert feedback and
LLM critics are used to iteratively revise an explicit natural-language
rare disease diagnostic procedure, hereafter referred to as a policy.
The resulting system, liteOdyssey, ranked the correct disease first in
59.3\% of 1,243 public benchmark cases spanning 722 rare diseases,
compared to 26.5\% for the corresponding baseline, with nearly identical
gains on the 679 diseases excluded from policy development (59.0\%
versus 25.9\% baseline). liteOdyssey was competitive with published
state-of-the-art rare-disease systems at less than one-thousandth the
deployment footprint. The same policy transferred across closed- and
open-weight language models of different sizes without retraining. After
adaptation to the Undiagnosed Diseases Network, it also improved
diagnostic accuracy in independent, hold-out real-world patients. Our
findings demonstrate that expert diagnostic reasoning can be
externalized into an inspectable and revisable policy that improves
diagnostic performance beyond the cases and diseases used to develop it
and transfers across model backbones.

\subsection{Results}\label{results}

\subsubsection{System overview}\label{system-overview}

liteOdyssey is an agentic diagnostic system for phenotype-first rare
disease diagnosis, built on a design principle: the diagnostic reasoning
procedure is externalized as an explicit natural-language policy rather
than encoded in model weights or orchestration code (Fig. 1a). Given a
patient's clinical features, as free text or as Human Phenotype Ontology
terms, an off-the-shelf LLM executes the policy end to end, drawing on a
library of specialized deterministic biomedical tools that return
source-traceable evidence from public knowledge bases. The output is a
ranked differential diagnosis in which each candidate carries its
associated gene, supporting evidence and confidence classification,
together with a continuous reasoning trace recording tool calls,
intermediate rankings, and reflective adjudication.

\begin{figure}
\centering
\includegraphics[width=0.85\linewidth,height=0.74\textheight,alt={Figure 1. liteOdyssey and how its policy is developed. (a) liteOdyssey guides an off-the-shelf reasoning LLM through phenotype interpretation, biomedical tool use, evidence review, and ranked differential diagnosis under an expert-guided diagnostic policy. Cases are described by HPO terms and narrative clinical text; the toolkit shown is the one used for the public benchmark cases. (b) The policy is developed through Policy Iteration with Human Feedback: an initial policy transcribed from expert clinical practice is executed by an off-the-shelf model on development cases, scored against benchmark outcomes and reasoning-process review, with failures mapped to implicated policy rules and proposed corrections expressed as concrete policy revisions. Revisions are admitted only when an expert judges the proposal clinically sound and a full-panel shows no regression. When development performance plateaus, the policy is frozen as a versioned, inspectable artifact; the model's weights are never updated. (c) Structural correspondence between reinforcement-learning post-training and PIHF across optimized object, reward model, human feedback, update operator, stability mechanism, data appetite, and consolidation; the correspondence is one of design, not formal equivalence.}]{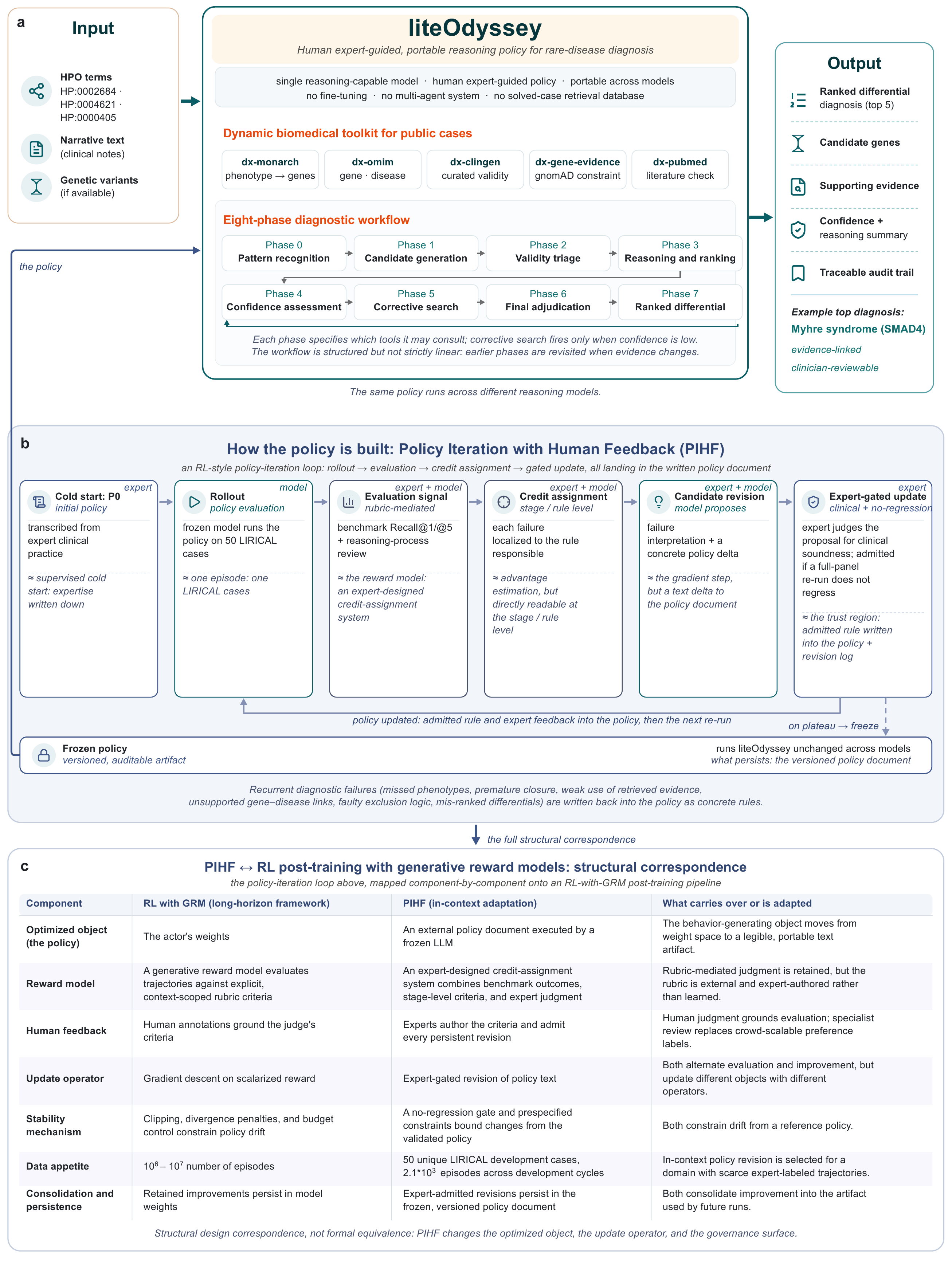}
\caption{Figure 1. liteOdyssey and how its policy is developed. (a)
liteOdyssey guides an off-the-shelf reasoning LLM through phenotype
interpretation, biomedical tool use, evidence review, and ranked
differential diagnosis under an expert-guided diagnostic policy. Cases
are described by HPO terms and narrative clinical text; the toolkit
shown is the one used for the public benchmark cases. (b) The policy is
developed through Policy Iteration with Human Feedback: an initial
policy transcribed from expert clinical practice is executed by an
off-the-shelf model on development cases, scored against benchmark
outcomes and reasoning-process review, with failures mapped to
implicated policy rules and proposed corrections expressed as concrete
policy revisions. Revisions are admitted only when an expert judges the
proposal clinically sound and a full-panel shows no regression. When
development performance plateaus, the policy is frozen as a versioned,
inspectable artifact; the model's weights are never updated. (c)
Structural correspondence between reinforcement-learning post-training
and PIHF across optimized object, reward model, human feedback, update
operator, stability mechanism, data appetite, and consolidation; the
correspondence is one of design, not formal equivalence.}
\end{figure}

The policy is a natural-language document that integrates expert
diagnostic reasoning principles, specialized biomedical tools, and rules
governing their use. It is organized into an eight-phase diagnostic
workflow (Fig. 1b). We developed the policy using Policy Iteration with
Human Feedback (PIHF), an in-context learning method. Starting from an
expert practice workflow as the initial policy, in each iteration an
off-the-shelf model runs the current policy once for each of the
development cases, a second model critiques each run using an
expert-designed credit-assignment system to locate where the reasoning
broke down, and clinicians decide which proposed revisions are
incorporated into the policy. PIHF therefore preserves the
evaluate-and-improve skeleton of reinforcement learning while changing
both the object being optimized and the update mechanism: an
expert-authored rubric replaces a learned reward model, and
clinician-gated edits to the natural-language policy replace
gradient-based weight updates (Fig. 1b and Fig. 1c). The natural
language policy, rather than the model weights, is the object
iteratively updated during development.

Given a case, the liteOdyssey system forms an initial impression
anchored on the most specific clinical features. Then, it generates
candidate genes from phenotype evidence and triages them by curated
gene--disease validity, progressing to build the case for and against
each candidate disease. When confidence is low or a feature cluster
remains unexplained, the policy directs corrective searches for recently
described or poorly indexed diseases, and a final reflective
adjudication re-examines the leading candidate against rarer but more
phenotypically specific competitors before the ranked differential is
issued. The workflow is structured but not strictly linear: the model
revisits earlier phases when new evidence changes the working
differential, and each phase specifies which tools it may consult, so
the tool library and the reasoning workflow form a single readable
artifact. The full phase-by-phase specification is provided in Methods
and Supplementary Information.

\subsubsection{Evaluation settings}\label{evaluation-settings}

We evaluated liteOdyssey on two public monogenic rare disease benchmarks
(Appendix Table A1): LIRICAL (370 cases across 252 diseases;
approximately 45\% ultra-rare)\citep{nguyen_diagnostic_2026} and the
PhenoPacket Store (873 cases, including a 300-case mapped/unmapped
stratified sample spanning 264 unique diseases and a 573-case extension
spanning 297 unique diseases; 52.8\% ultra-rare). Unmapped causal
diseases could not be linked to Orphanet and served as a proxy for
recently described or poorly indexed diseases (Methods). Together, the
benchmarks comprised 1,243 cases spanning 722 diseases.

To test whether PIHF produces a reusable diagnostic policy that
generalizes beyond the cases and diseases used to develop it, policy
development was strictly separated from evaluation. Only 50 LIRICAL
cases, and no PhenoPacket Store cases, were used for policy development;
the remaining 1,193 cases, spanning 679 diseases that never appeared
during development, constituted an independent hold-out set. To assess
generalizability to a real-world patient cohort, we further evaluated
the system across 515 patients from the Undiagnosed Diseases Network
(UDN), whose diagnoses remained unresolved after standard clinical
workup and were subsequently established by the network, comprising 447
distinct diagnoses across 15 U.S. clinical sites. We applied PIHF to 50
UDN cases to adapt the policy to atypical clinical presentations and
held out the remaining 465 patients for independent evaluation
(Methods).

The primary comparator throughout is each model's own parametric
baseline: the same model but without the diagnostic policy, biomedical
tools, or external information sources, answering from pre-trained
knowledge alone. A cumulative ablation ladder further isolates the
contributions of optimized prompting and open-ended agentic source
access from the full policy (Methods). For direct comparability with
published systems, outputs were scored with the standardized
LLM-as-a-judge protocol and cross-checked against deterministic
exact-OMIM identifier matching. To mitigate potential same-family judge
bias, all comparisons were additionally re-scored under the same
protocol using an independent judge from another model family (Methods).

In the UDN cohort, we further validated these automated scores against
blinded physician adjudication (Methods).

\subsubsection{Gains across public
benchmarks}\label{gains-across-public-benchmarks}

Across 1,243 public benchmark cases, liteOdyssey ranked the correct
disease first in 59.3\% of cases, versus 26.5\% for the parametric
baseline (Fig. 2a).

\begin{figure}
\centering
\includegraphics[width=0.88\linewidth,height=0.72\textheight,alt={Figure 2. Policy-guided reasoning improves disease ranking across public rare disease benchmarks. (a) Pooled disease Recall@1 on the same 1,243 public benchmark cases: parametric baselines for the four execution models (grey) and the liteOdyssey full system with its GPT-5.4 backbone (blue; 737 of 1,243 cases). (b--e) Disease Recall@1 and Recall@5 for the parametric baseline and the full liteOdyssey system across four public benchmark settings: (b) LIRICAL, (c) PhenoPacket Store mapped cases, (d) PhenoPacket Store unmapped cases and (e) PhenoPacket Store extension. Recall@1 is the proportion of cases in which the correct diagnosis was ranked first; Recall@5 is the proportion in which it appeared within the top five. The parametric baseline used the same language model to diagnose from the same clinical features without the liteOdyssey diagnostic policy or biomedical tools; the full system used the PIHF-derived policy-guided workflow with biomedical tool use. Outputs were scored with an LLM-as-a-judge protocol following prior work, with judge and diagnostic system always from different model families. A deterministic exact-OMIM identifier cross-check was applied throughout (Methods).}]{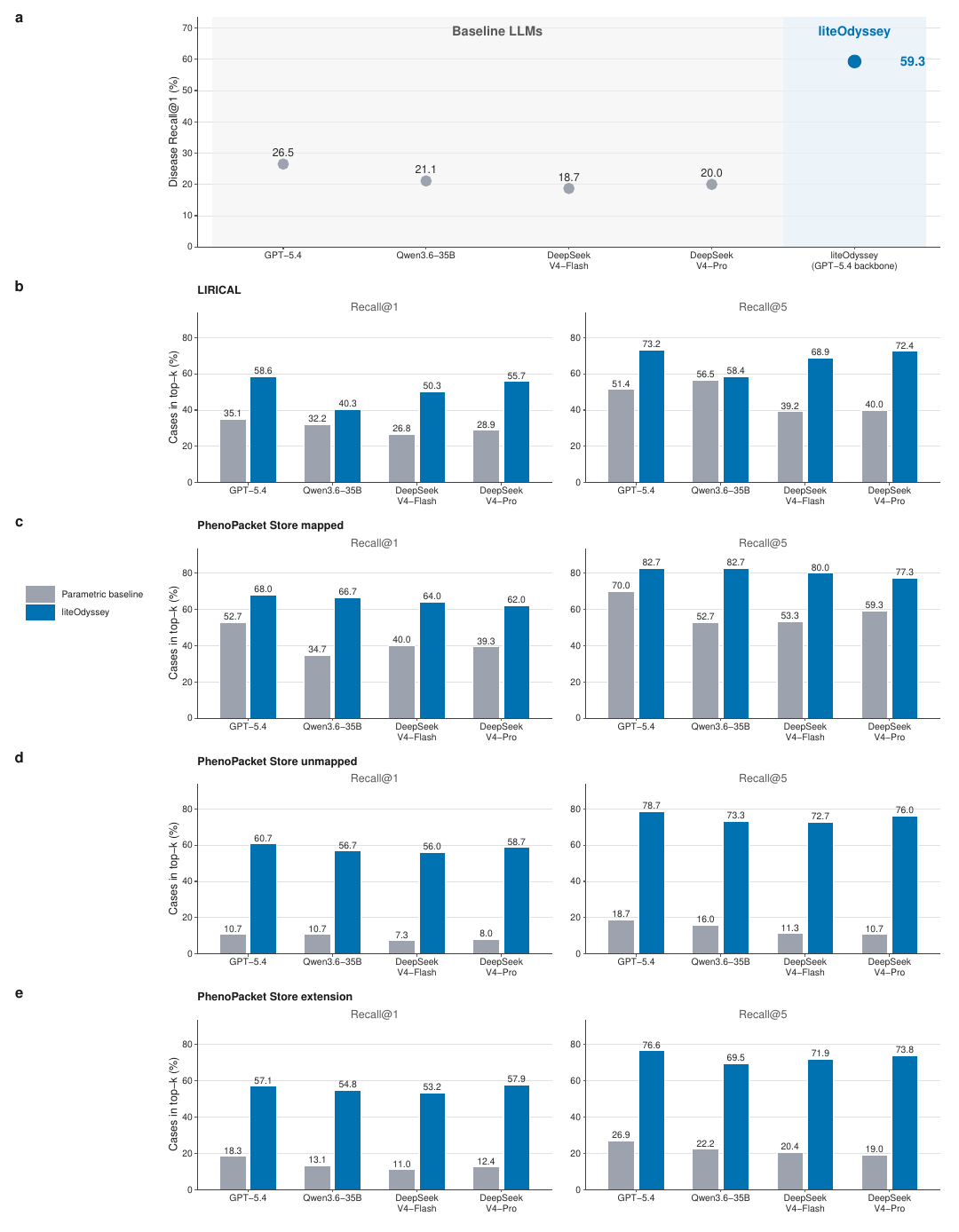}
\caption{Figure 2. Policy-guided reasoning improves disease ranking
across public rare disease benchmarks. (a) Pooled disease Recall@1 on
the same 1,243 public benchmark cases: parametric baselines for the four
execution models (grey) and the liteOdyssey full system with its GPT-5.4
backbone (blue; 737 of 1,243 cases). (b--e) Disease Recall@1 and
Recall@5 for the parametric baseline and the full liteOdyssey system
across four public benchmark settings: (b) LIRICAL, (c) PhenoPacket
Store mapped cases, (d) PhenoPacket Store unmapped cases and (e)
PhenoPacket Store extension. Recall@1 is the proportion of cases in
which the correct diagnosis was ranked first; Recall@5 is the proportion
in which it appeared within the top five. The parametric baseline used
the same language model to diagnose from the same clinical features
without the liteOdyssey diagnostic policy or biomedical tools; the full
system used the PIHF-derived policy-guided workflow with biomedical tool
use. Outputs were scored with an LLM-as-a-judge protocol following prior
work, with judge and diagnostic system always from different model
families. A deterministic exact-OMIM identifier cross-check was applied
throughout (Methods).}
\end{figure}

The performance improvement was consistent across both benchmarks:
58.6\% versus 35.1\% on LIRICAL, and 59.6\% versus 22.9\% on the
PhenoPacket Store.

The largest gains occurred in cases for which the diagnosis was less
likely to be represented in the model's pre-trained knowledge (Fig. 2d).
Among PhenoPacket Store cases whose causal disease could not be linked
to Orphanet disease identifiers, a proxy for recently described or
poorly indexed conditions, Recall@1 increased from 10.7\% at baseline to
60.7\% with liteOdyssey, and Recall@5 from 18.7\% to 78.7\%. These
results suggest that liteOdyssey can identify diagnoses that are not
readily recovered from the model's pre-trained knowledge alone by
systematically seeking external evidence and weighing it against the
patient's presentation.

On the LIRICAL benchmark, our results compare favorably with the
strongest published reasoning-based system: DeepRare reported 39.5\%
Recall@1 without retrieval from its solved-case database and
51.6--56.0\% with retrieval from 67,795 solved cases, whereas
liteOdyssey achieved 58.6\% on the same benchmark using a single
reasoning model, public biomedical tools and no solved-case
database\citep{zhao_agentic_2026}. Taken together, a policy developed
from 50 cases generalized to substantially improve Recall@1 across 1,243
cases and 722 diseases (26.5\% to 59.3\%) and placed the correct
diagnosis in the top five for 73.2\% of LIRICAL and 78.0\% of
PhenoPacket Store cases.

\subsubsection{Policy generalizes to unseen
diseases}\label{policy-generalizes-to-unseen-diseases}

If the policy merely steered the model toward knowledge it already
possessed, its gains should diminish on diseases never encountered
during policy development. Policy development used only 50 LIRICAL
cases, spanning 43 diseases, and no PhenoPacket Store cases. The
remaining LIRICAL cases therefore provide a case-level hold-out, whereas
PhenoPacket Store provides an entirely independent, development-excluded
benchmark. Of the 43 development diseases, 12 recurred as different
cases among the 320 hold-out LIRICAL cases. Of the 722 diseases across
the two benchmarks, 679 never appeared during development; these span
619 distinct causal genes, of which 614 (99.2\%) were absent from the 46
causal genes represented among the 43 development diseases, and the
remaining five were causal for different diseases in the two sets.

Gains on the hold-out cases were nearly identical to those on the full
cohort. On the 320 hold-out LIRICAL cases, liteOdyssey reached 54.4\%
Recall@1 against 33.1\% for the parametric baseline, and 71.2\% versus
47.8\% at Recall@5, corresponding to gains of 21.2 and 23.4 percentage
points, respectively, compared with gains of 21.6 and 24.1 percentage
points on the full cohort (Fig. 3a--c).

\begin{figure}
\centering
\includegraphics[width=1\linewidth,height=\textheight,keepaspectratio,alt={Figure 3. Policy gains persist on development-excluded cases. Held-out analyses exclude the 50 cases used during policy development. (a) Disease Recall@1 and Recall@5 on 320 development-excluded LIRICAL cases. (b) Disease Recall@1 and Recall@5 on 465 development-excluded Undiagnosed Diseases Network (UDN) cases. The policy was executed by GPT-5.4 for LIRICAL and GPT-5.3 for UDN. Bars compare the matched parametric baseline with the full liteOdyssey system; paired exact McNemar p-values for the UDN cohort. (c) Full-system-minus-baseline improvement for the full cohort and development-excluded set (labels mark development-excluded values), showing that gains were preserved after exclusion of development cases. All panels report the deterministic exact-OMIM identifier cross-check (Methods); LLM-as-a-judge values, reported for comparability with published systems, accompany the public-benchmark and UDN results (Figs. 2 and 4).}]{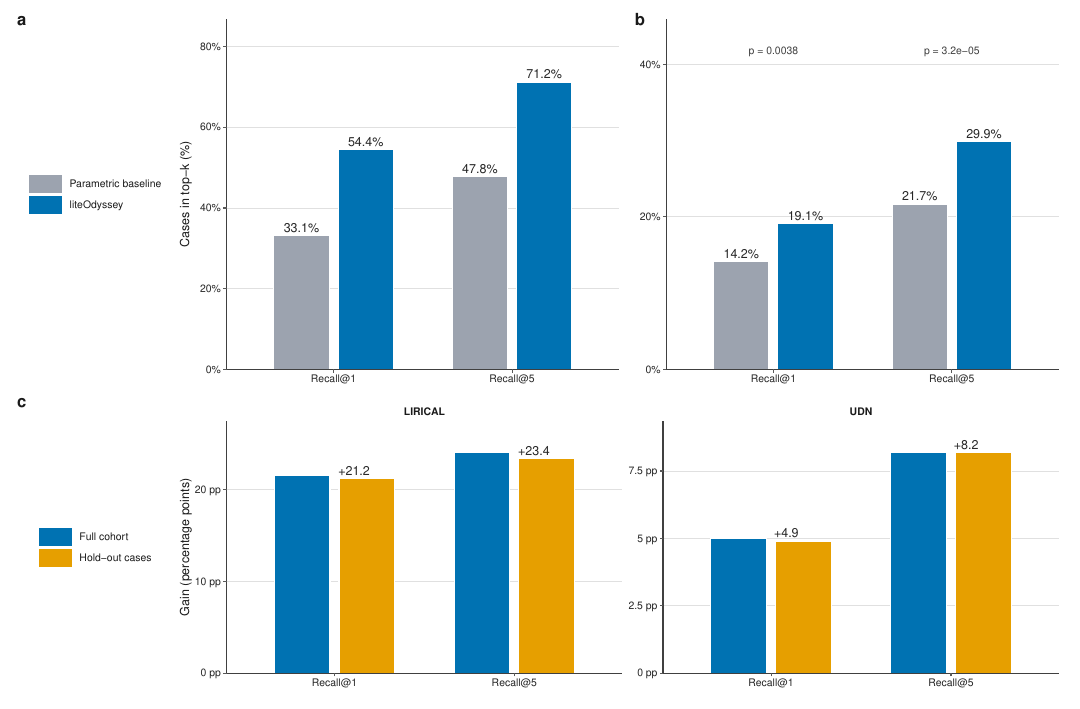}
\caption{Figure 3. Policy gains persist on development-excluded cases.
Held-out analyses exclude the 50 cases used during policy development.
(a) Disease Recall@1 and Recall@5 on 320 development-excluded LIRICAL
cases. (b) Disease Recall@1 and Recall@5 on 465 development-excluded
Undiagnosed Diseases Network (UDN) cases. The policy was executed by
GPT-5.4 for LIRICAL and GPT-5.3 for UDN. Bars compare the matched
parametric baseline with the full liteOdyssey system; paired exact
McNemar p-values for the UDN cohort. (c) Full-system-minus-baseline
improvement for the full cohort and development-excluded set (labels
mark development-excluded values), showing that gains were preserved
after exclusion of development cases. All panels report the
deterministic exact-OMIM identifier cross-check (Methods);
LLM-as-a-judge values, reported for comparability with published
systems, accompany the public-benchmark and UDN results (Figs. 2 and
4).}
\end{figure}

Because no PhenoPacket Store cases were used during policy development,
and 679 of the 722 diseases across the public benchmarks were absent
from development, the combined evaluation provides a test of whether the
policy transfers beyond the diseases from which it was constructed. The
preservation of performance gains on development-excluded diseases
argues against the improvement arising primarily from case- or
disease-specific reasoning strategies acquired during PIHF. The limited
overlap in causal genes further suggests that transfer was not confined
to alternative phenotypes of genes encountered during development.
Together, these results support the conclusion that PIHF produces
reusable diagnostic reasoning principles that can be applied to diseases
outside the policy development set, rather than simply accumulating
solutions to the diseases encountered during development. This
distinction is particularly relevant in rare disease diagnosis, where
successful reasoning must often extend beyond familiar or typical
clinical presentations.

\subsubsection{Policy generalizes across
models}\label{policy-generalizes-across-models}

The policy transferred across models as an agent skill, i.e., an
external natural-language reasoning procedure executed by a
general-purpose model during inference. Developed with a frontier model
(GPT-5.4), the policy was provided unmodified to open-weight
Mixture-of-Experts models spanning a sixteen-fold range of active
parameters: Qwen3.6-35B-A3B (3 billion active),
DeepSeek-V4-Flash-Preview (13 billion active), and
DeepSeek-V4-Pro-Preview (49 billion active). Every model improved over
its own parametric baseline (Fig. 2a). This consistency is notable
because recent agent-skill studies have found that the utility of
external skills can vary substantially across tasks and executors: some
skills yield gains of 4 to 26 percentage points but degrade on 13 of 87
tasks, model-generated skills show no significant improvement across 56
tasks, and retrieval can mislead agents; no study to date isolates the
skill's effect from the scale of its
executor\citep{li_skillsbench_2026, huang_llm_generated_2026, liu_how_2026}.
In contrast, the same liteOdyssey policy improved diagnostic performance
across every model tested regardless of size, a notable result against
the scaling-law expectation that capability improves with model
scale\citep{kaplan_scaling_2020, hoffmann_training_2022}. The
3-billion-active-parameter model reached 57.2\% Recall@1 on the
PhenoPacket Store, within 2.4 points of the frontier model (59.6\%),
despite substantially lower parametric baseline performance of
10.7--34.7\%. Larger open-weight models approached the frontier model on
both benchmarks. Policy transfer also extended to a second closed-weight
model family: Claude Opus 4.6 reached 57.6\% R@1 on LIRICAL and 63.3\%
and 58.0\% on the mapped and unmapped PhenoPacket cohorts (Appendix
Figure A1). Thus, the diagnostic reasoning strategies learned through
PIHF were not specific to the model used to develop it and could be
executed effectively by models differing in architecture and size.

However, we note that model size nevertheless remained consequential.
The smallest model sometimes entered circular reasoning loops and failed
to terminate within the compute limit, requiring additional attempts
that were triggered by non-completion, and a substantial gap remained on
LIRICAL (40.3\% versus 58.6\% for the frontier model). The policy
therefore did not eliminate differences in underlying model capability.
Rather, its consistent improvement across model sizes and families
supports its interpretation as a portable reasoning layer rather than a
model-specific optimization.

liteOdyssey also generalized at the causal-gene level: on the
PhenoPacket Store, the full system ranked the causal gene first in
61.2\% of cases versus 19.2\% for the parametric baseline, and the
smallest open-weight model (Qwen3.6-35B) showed the same pattern (60.0\%
versus 8.4\%) (Appendix Figure A2).

\subsubsection{Policy adapts to a real-world clinical cohort and
improves diagnostic
performance}\label{policy-adapts-to-a-real-world-clinical-cohort-and-improves-diagnostic-performance}

While public benchmarks serve as useful controlled tests, they do not
necessarily capture the diagnostic complexity of real-world rare disease
patients. Using GPT-5.3, we evaluated liteOdyssey in a cohort of 515
Undiagnosed Diseases Network (UDN) patients.

For the UDN cohort, we used PIHF to extend the policy developed on the
public benchmarks (reproduced in Appendix Figure B2), adding targeted
reasoning steps for the atypical and multi-system presentations often
seen among UDN patients. Fifty cases were used for this adaptation and
the remaining 465, spanning 407 diagnoses, were held out from policy
development. On the full UDN cohort, liteOdyssey achieved a Recall@1 of
20.4\%, compared with 16.7\% for the parametric baseline (+3.7
percentage points; paired exact McNemar p = 0.027; 95\% CI, 0.6 to 6.8),
and a Recall@5 of 31.8\% versus 25.6\% for the baseline (+6.2 percentage
points; p = 7 × 10\textsuperscript{-4}) (Fig. 4a). The gains persisted
in the 465 development-excluded cases: Recall@1 was 19.1\% versus 15.5\%
for the parametric baseline, and Recall@5 was 29.9\% versus 23.7\% (Fig.
3b, c).

\begin{figure}
\centering
\includegraphics[width=1\linewidth,height=\textheight,keepaspectratio,alt={Figure 4. Policy-guided reasoning improves diagnosis in a real-world UDN cohort, corroborated by blinded physician review. (a) Disease Recall@1 and Recall@5 for 515 patients evaluated by the Undiagnosed Diseases Network (UDN). The full liteOdyssey system is compared with a matched parametric baseline using the same model and clinical features without the policy or biomedical tools. (b) Distribution of physician-adjudicated differential-diagnosis scores (Bond scale: 0, no close suggestion, to 5, exact diagnosis) for the same two arms, weighted to the full 515-case cohort. Two blinded physicians scored a prespecified 120-case sample and a third adjudicated disagreements; error bars are pointwise 95\% Wilson intervals. liteOdyssey increased Recall@1 from 16.7\% to 20.4\% and Recall@5 from 25.6\% to 31.8\%; under physician review it scored the exact diagnosis in 31.8\% of outputs versus 26.6\% baseline and had fewer complete misses (12.7\% versus 15.9\%), with a higher weighted mean score (3.15 versus 2.91; paired difference +0.24, 95\% CI +0.09 to +0.40). Recall@1 and Recall@5 are the proportions of cases with the correct diagnosis ranked first or within the top five. Outputs in a were scored with an LLM-as-a-judge protocol following prior work, with judge and diagnostic system always from different model families.}]{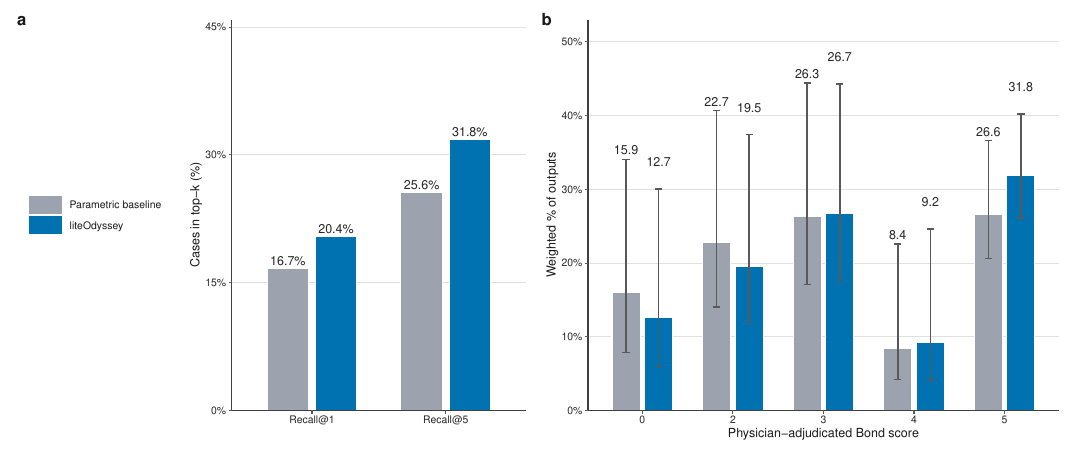}
\caption{Figure 4. Policy-guided reasoning improves diagnosis in a
real-world UDN cohort, corroborated by blinded physician review. (a)
Disease Recall@1 and Recall@5 for 515 patients evaluated by the
Undiagnosed Diseases Network (UDN). The full liteOdyssey system is
compared with a matched parametric baseline using the same model and
clinical features without the policy or biomedical tools. (b)
Distribution of physician-adjudicated differential-diagnosis scores
(Bond scale: 0, no close suggestion, to 5, exact diagnosis) for the same
two arms, weighted to the full 515-case cohort. Two blinded physicians
scored a prespecified 120-case sample and a third adjudicated
disagreements; error bars are pointwise 95\% Wilson intervals.
liteOdyssey increased Recall@1 from 16.7\% to 20.4\% and Recall@5 from
25.6\% to 31.8\%; under physician review it scored the exact diagnosis
in 31.8\% of outputs versus 26.6\% baseline and had fewer complete
misses (12.7\% versus 15.9\%), with a higher weighted mean score (3.15
versus 2.91; paired difference +0.24, 95\% CI +0.09 to +0.40). Recall@1
and Recall@5 are the proportions of cases with the correct diagnosis
ranked first or within the top five. Outputs in a were scored with an
LLM-as-a-judge protocol following prior work, with judge and diagnostic
system always from different model families.}
\end{figure}

Absolute differential diagnostic accuracy was lower in the UDN cohort
than in the public benchmarks, consistent with the greater diagnostic
complexity of patients referred to the UDN. By comparison, off-the-shelf
LLMs previously evaluated on UDN patients achieved physician-confirmed
exact-diagnosis rates of 13.3\% (ChatGPT-4o) across 90
cases\citep{shyr_large_2025}. Importantly, the magnitude of
liteOdyssey's improvement was similar after exclusion of the 50 cases
used for UDN-specific policy adaptation. These results show that PIHF
can adapt a policy developed on curated benchmark cases to a
substantially different clinical setting using limited local development
data, with the resulting gains persisting in hold-out, real-world
patients.

To test whether the improvements measured by automated scoring were also
supported by physician adjudication, two physicians independently
adjudicated a prespecified 120-case validation sample, blinded to model
identity, scoring each output on the Bond differential-diagnosis
scale\citep{bond_differential_2012}; a third blinded physician resolved
the 66 output-level disagreements. On physician-adjudicated scores
weighted to the full 515-case cohort, liteOdyssey scored higher than the
parametric baseline (weighted mean 3.15 versus 2.91; paired difference
+0.24, 95\% CI +0.09 to +0.40; two-sided p = 0.002). liteOdyssey
produced the exact diagnosis for 31.8\% of patients versus 26.6\% for
the baseline, exact or very close matches for 41.1\% versus 35.0\%, and
complete misses for 12.7\% versus 15.9\% (Fig. 4b); inter-rater
agreement was high (quadratic-weighted κ = 0.909). Physician agreement
with the LLM judge was strongest on high-confidence outputs (86--87\%)
and lower on moderate- and low-confidence outputs, indicating that the
judge's reported confidence can stratify the reliability of automated
scoring.

\subsubsection{Isolating the policy's contribution beyond prompting and
access to external
information}\label{isolating-the-policys-contribution-beyond-prompting-and-access-to-external-information}

Using GPT-5.4 and exact OMIM identifier matching, we performed ablation
studies to isolate liteOdyssey's performance gain beyond prompt
engineering or biomedical resource access alone. On LIRICAL, Recall@1
increased from 34.3\% with the base prompt to 36.8\% with a prompt
automatically optimized by GPT-5.5, 44.9\% with access to PIHF-optimized
public biomedical resources, and 55.9\% with the full policy (Fig. 5;
Appendix Figure C1 and Appendix Table A2). Recall@5 showed the same
pattern (48.6\%, 52.7\%, 61.9\%, and 72.7\%).

\begin{figure}
\centering
\includegraphics[width=1\linewidth,height=\textheight,keepaspectratio,alt={Figure 5. The policy provides additional performance gains beyond optimized prompting and PIHF-optimized resource access across three benchmark cohorts. Cumulative ablation on LIRICAL (370 cases), PhenoPacket Store unmapped (150 cases) and PhenoPacket Store extension (573 cases), all executed on GPT-5.4 and scored by the deterministic exact-OMIM identifier cross-check (Methods): the base model given clinical features only; an optimized diagnostic prompt without tools; that prompt with access to the same PIHF-optimized public biomedical sources; and the full liteOdyssey system with the diagnostic policy. Rungs 1 and 4 are the GPT-5.4 benchmark runs of Fig. 2, scored here by exact OMIM identifier matching. Performance of the parametric baseline was low on the PhenoPacket Store cohorts, whose diseases are less likely to be represented in the model's pre-trained knowledge, and optimized prompting alone does not recover it. Access to the PIHF-optimized resources closes part of the gap, with additional gains from the full policy. The complete grid, including development-excluded slices, is reported in the supplement (Appendix Table A2).}]{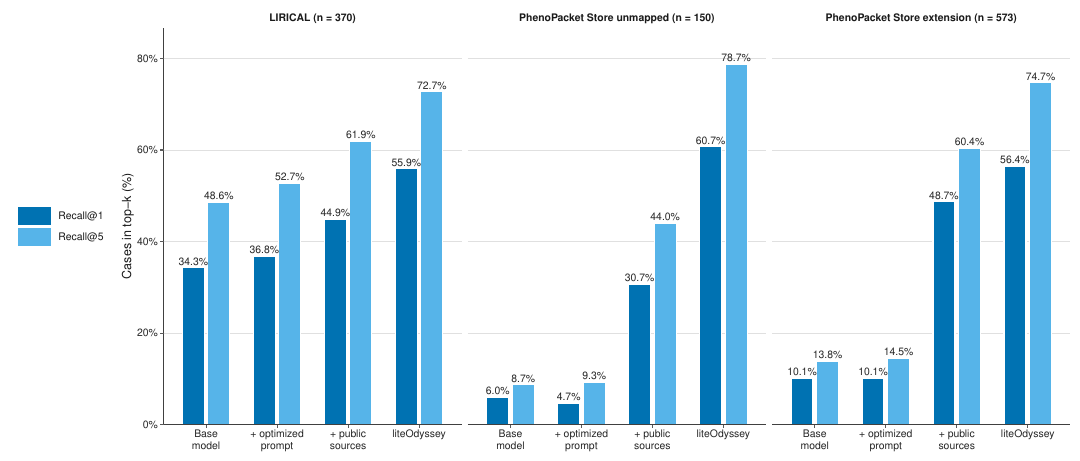}
\caption{Figure 5. The policy provides additional performance gains
beyond optimized prompting and PIHF-optimized resource access across
three benchmark cohorts. Cumulative ablation on LIRICAL (370 cases),
PhenoPacket Store unmapped (150 cases) and PhenoPacket Store extension
(573 cases), all executed on GPT-5.4 and scored by the deterministic
exact-OMIM identifier cross-check (Methods): the base model given
clinical features only; an optimized diagnostic prompt without tools;
that prompt with access to the same PIHF-optimized public biomedical
sources; and the full liteOdyssey system with the diagnostic policy.
Rungs 1 and 4 are the GPT-5.4 benchmark runs of Fig. 2, scored here by
exact OMIM identifier matching. Performance of the parametric baseline
was low on the PhenoPacket Store cohorts, whose diseases are less likely
to be represented in the model's pre-trained knowledge, and optimized
prompting alone does not recover it. Access to the PIHF-optimized
resources closes part of the gap, with additional gains from the full
policy. The complete grid, including development-excluded slices, is
reported in the supplement (Appendix Table A2).}
\end{figure}

The PhenoPacket Store cohorts showed the same ordered pattern across the
four configurations: 6.0\%, 4.7\%, 30.7\%, and 60.7\% Recall@1 on the
unmapped cohort (150 cases), and 10.1\%, 10.1\%, 48.7\%, and 56.4\% on
the extension cohort (573 cases) (Fig. 5).

These results separate three successive levels of support provided to
the same underlying model. Optimizing the diagnostic prompt produced
little improvement across the PhenoPacket Store cases and only a modest
improvement on LIRICAL, whereas access to biomedical resources produced
an intermediate improvement. The largest performance gain was observed
when the model was guided by the policy. Therefore, the policy
contributes beyond giving the model more instructions or more
information; it provides a structured procedure for gathering, weighing,
and revisiting evidence during rare disease diagnosis.

\subsection{Discussion}\label{discussion}

We showed that expert rare disease reasoning can be externalized as an
inspectable and transferrable diagnostic policy that improves language
models across three distinct axes of generalization: diseases excluded
from policy development, model families and sizes, and, after limited
local adaptation, hold-out patients from the Undiagnosed Diseases
Network. The method behind this generalization is PIHF, which converts
expert review and model critique of reasoning failures into revisions of
a persistent policy. Developed from 50 cases, the policy more than
doubled Recall@1 across 1,243 cases (26.5\% to 59.3\%), placed the
correct diagnosis among the top five in more than 70\% of cases across
both benchmarks, retained nearly identical improvements across 679
diseases held-out from development, and improved every model backbone
tested. Clinicians can inspect, revise, and govern the resulting policy
directly, because its diagnostic logic lives in explicit
natural-language reasoning principles, specialized biomedical tools, and
rules governing their use rather than in opaque model weights.

The patterns of these gains indicate that PIHF captured a generalizable
diagnostic reasoning procedure for rare disease. Had the policy
accumulated disease-specific strategies, improvement should have
diminished outside the 50 development cases; instead, it persisted
across 679 development-excluded diseases, including 614 causal genes not
represented among them. Had the policy encoded model-specific behavior,
transfer should have failed; instead, the unchanged policy improved
every tested model across open- and closed-weight families and a
sixteen-fold range of active parameters. Model capability nevertheless
remained important: smaller models executed the procedure less reliably
on some benchmarks and occasionally failed to terminate. The policy
therefore did not replace model capability but added a reusable
reasoning layer that transferred across heterogeneous backbones.

The ablation experiments further distinguish structured reasoning from
better prompting or access to external information alone. Optimizing the
diagnostic prompt produced little or modest improvement, whereas access
to the PIHF-optimized set of public biomedical resources accounted for a
substantial portion of the gain; a randomly chosen source set might have
delivered less. Performance increased further when the complete
PIHF-derived policy was used, including its specialized tools and rules
governing evidence retrieval and adjudication. This gain shows that
information access alone was insufficient to reproduce liteOdyssey's
performance. This is particularly relevant for rare disease diagnosis,
where the challenge is not only retrieving disease knowledge, but also
determining which evidence is diagnostically informative, when
discordant findings should be discounted, and how competing explanations
should be integrated into a diagnosis.

PIHF is the learning method; liteOdyssey is the system it produced. In
PIHF, a frozen model executes the current policy, expert-designed credit
assignment localizes recurrent failures, and clinicians admit clinically
sound, reusable revisions to a versioned policy. liteOdyssey combines
the resulting reasoning workflow, clinician-vetted criteria, biomedical
tools, and rules for using them. These components constitute the policy
and were iteratively developed rather than assembled post hoc. The tools
and the criteria governing their use co-evolved with the reasoning
procedure. The resulting policy is therefore not simply a prompt
provided to a language model, but an executable representation of expert
diagnostic reasoning in natural language that explicitly defines what
evidence to seek, how to obtain it, how to weigh competing findings, and
when to revisit an earlier hypothesis.

In rare disease diagnosis, the correct answer is generally unavailable
while the task is being performed, limiting approaches that require an
objective or reward signal during inference. What can be improved is the
process of deriving the answer: when the reasoning is sound, the correct
diagnosis follows as its byproduct, whereas a correct label reached by
unsound reasoning gives clinicians and patients little to trust or act
on. Prompt optimizers, self-reflection methods, and in-context
reinforcement learning generally presuppose an available answer or
scoring metric, and their lessons remain tied to the tasks or attempts
that produced
them\citep{khattab_dspy_2024, agrawal_gepa_2025, yuksekgonul_optimizing_2025, shinn_reflexion_2023, song_reward_2025}.
PIHF instead consolidates accepted lessons into a persistent, versioned
policy revised through expert assessment of the reasoning process
(Methods). The expert gate makes this possible without a
machine-readable objective: clinicians specify what evidence to seek,
which actions to take, and how reasoning should be judged, then
adjudicate revisions at gated checkpoints so feedback becomes reusable
guidance for future cases.

The UDN evaluation illustrates both the promise and the boundary of this
generalization. The benchmark-derived policy was further adapted with 50
UDN cases to incorporate the atypical presentations often encountered in
this clinical setting. The resulting performance gains persisted in the
remaining 465 development-excluded UDN cases and were confirmed by
blinded physician adjudication. This suggests that a reusable reasoning
policy can provide a starting point for clinical deployment while still
benefiting from limited local adaptation, analogous to adapting other
clinical systems to differences in population and workflow. Prospective
evaluation is needed to determine how much adaptation is needed across
institutions, disease populations, and clinical tasks.

Future work should prospectively evaluate safety, efficiency, cost,
clinician trust, patient outcomes, and the extent to which individual
policy components transfer unchanged or require local revision.
Stabilized policies might also be distilled into model weights, trading
auditability and cross-model portability for lower inference
cost\citep{snell_learning_2022, ye_policy_2026, charakorn_doc_lora_2026}.

liteOdyssey shows that human expertise can become a scalable AI
capability through governed learning rather than model training alone.
By converting recurrent failures and expert corrections into an explicit
reasoning policy that transfers beyond its development diseases and
across model backbones, PIHF provides a mechanism for making scarce
diagnostic expertise reusable without embedding it in opaque model
weights. For rare disease diagnosis, where clinician expertise is scarce
and the knowledge landscape continually evolves, this offers a path
toward AI systems whose reasoning can improve over time while remaining
under human oversight.

\subsection{Online Methods}\label{online-methods}

\subsubsection{Experimental design}\label{experimental-design}

The design tested four properties of a PIHF-derived natural-language
policy: improvement over each model's parametric baseline,
generalization to diseases absent from development, transfer across
models, and performance in a real-world clinical cohort. The primary
endpoint was disease Recall@1 under the standardized LLM-as-a-judge
protocol; Recall@5 was secondary, with a deterministic exact-OMIM
cross-check throughout (Methods, `Outcomes and scoring'; Appendix Table
A3). Prespecified components were: (i) development restricted to 50
LIRICAL cases and, for the UDN extension, 50 UDN cases, with all others
held out (Methods, `Benchmarks and clinical cohort'); (ii) paired
full-system versus parametric-baseline comparisons in every cohort;
(iii) a four-configuration cumulative ablation on one model under
exact-OMIM scoring (Methods, `Ablation, cross-model evaluation, and
reproducibility'); (iv) cross-model evaluation of the unchanged policy
across five models (Methods); and (v) blinded physician validation in a
prespecified, stratified 120-case UDN sample weighted to the full
cohort, with four prespecified adjudication sensitivity analyses
(Methods). Recall comparisons used exact McNemar tests with 95\%
confidence intervals from a 20,000-iteration bootstrap. The physician
comparison used a design-based Wald test, inter-rater agreement used
quadratic-weighted κ, weighted proportions used pointwise 95\% Wilson
intervals, and gene-level endpoints were Bonferroni-corrected across the
two recall metrics. All eligible cases were evaluated; no sample-size
calculation was performed.

\subsubsection{System overview and diagnostic
policy}\label{system-overview-and-diagnostic-policy}

liteOdyssey accepts patient clinical features as free text or Human
Phenotype Ontology {[}HPO{]} terms and returns an unstructured ranked
differential with candidate diseases, associated genes, supporting
evidence, and a traceable reasoning summary; evaluation converts this
output into categorical Recall@1 and Recall@5 outcomes. It comprises an
external natural-language policy, a biomedical tool library providing
structured access to curated public sources, and a reasoning-capable
model that executes the policy end to end. The policy defines how to
interpret features, select tools (Methods, `Biomedical tool library'),
weigh returned evidence, and revisit earlier steps, making the workflow
and tool library one readable artifact. Because the policy is external,
liteOdyssey requires no weight updates, solved-case database, or
multi-agent orchestration. The same specification can run across model
families, and clinicians can inspect, contest, or revise it without
retraining. During execution, the model integrates evidence across tool
calls into a continuous case-level reasoning process.

The policy developed by PIHF (Methods, `Policy Iteration with Human
Feedback') is an eight-phase reasoning workflow: (0) pattern
recognition, (1) phenotype-based candidate gene generation, (2) evidence
triage, (3) deep disease-level investigation, (4) confidence assessment,
(5) corrective search, (6) reflective adjudication, and (7) a ranked
final output. The workflow is structured but not strictly linear: the
system can revisit earlier phases when new evidence changes the working
differential or when uncertainty remains high, and each phase specifies
which biomedical tools (Methods) it draws on. The full phase-by-phase
specification is provided in Appendix B.1.

\subsubsection{Biomedical tool library}\label{biomedical-tool-library}

We describe the subset of tools exercised by the phenotype-first
evaluations. Each addresses a focused clinical question and returns
deterministic, source-traceable evidence from a public source; no
individual tool returns a differential diagnosis. The language model
interprets and integrates this evidence into the final ranking.

Five knowledge tools were evaluated: phenotype-to-gene ranking (Monarch
Initiative\citep{mungall_monarch_2016}); gene-to-disease and
disease-family lookup with inheritance and onset
(OMIM\citep{amberger_omimorg_2014}); gene-disease validity and dosage
sensitivity (ClinGen\citep{rehm_clingen_2015}); gene constraint
(gnomAD\citep{karczewski_mutational_2020}, combined with OMIM); and
literature retrieval (PubMed). The broader toolkit includes
variant-quality and inheritance-pattern components that activate when
sequencing or trio data are available, but these were not exercised on
the phenotype-only benchmarks. Appendix Table A4 specifies each
evaluated tool's clinical question, source, and returned evidence.
Together, the tools support phenotype-, gene-, and disease-level
reasoning.

\subsubsection{Policy Iteration with Human Feedback
(PIHF)}\label{policy-iteration-with-human-feedback-pihf}

PIHF adapts the evaluate-and-improve structure of generalized policy
iteration to an external natural-language
policy\citep{sutton_reinforcement_2018}. Rare disease diagnosis differs
from settings with programmatically verifiable objectives, such as
formal mathematical reasoning\citep{hubert_olympiad_level_2025}: the
final diagnosis can be confirmed retrospectively, but outcome-only
scoring cannot distinguish a sound diagnostic process from a lucky
answer or define how to interpret phenotypes, weigh evidence, revisit
hypotheses, and decide when evidence is sufficient. Human feedback can
construct objectives that programs cannot specify
directly\citep{christiano_deep_2017, ziegler_fine_tuning_2019}. Modern
reinforcement-learning pipelines similarly use rule-based verifiers for
machine-checkable behavior and generative reward models with explicit
rubrics for behavior that rules cannot
score\citep{deepseek_ai_deepseek_v4_2026, kimi_team_kimi_2026, hou_single_rollout_2026}.
In PIHF, an expert rubric defines the quality of diagnostic reasoning,
allowing bounded human judgment to govern many
rollouts\citep{liu_inference_time_2025}. Expert-labeled diagnostic
trajectories are not available at the scale required for weight-space
optimization, whereas pre-trained models can follow complex
natural-language task
specifications\citep{radford_language_2019, brown_language_2020}. An
off-the-shelf model therefore executes the policy, a second model
critiques the trajectory against the expert-designed credit-assignment
system, and clinician-gated revisions update the policy text rather than
model weights.

The mapping preserves each function under a blinded-reasoning
constraint: the true diagnosis remains hidden until evaluation, and all
loop operations use policy-visible evidence (Fig. 1b). An expert
workflow supplies the supervised cold start; an expert-authored rubric
serves the function of a learned reward model; the benchmark outcome
provides the verifiable anchor while expert judgment evaluates the
reasoning path. Stage-level attribution replaces advantage estimation:
because the policy is structured text, the critique can localize a
failure to the phase and rule that produced it. A no-regression gate
serves the function of a trust region, bounding drift against a clinical
criterion. Expert admission consolidates accepted changes into a
versioned document rather than a model checkpoint. This in-context
design avoids requiring a crowdsourced preference dataset large enough
to train a reward model. The correspondence is structural, not
algorithmic; PIHF provides neither a gradient estimator nor a formal
improvement guarantee. A detailed formal treatment of PIHF is available
in this paper\citep{nguyen_policy_2026}.

For each public-policy rollout, GPT-5.4 at high reasoning effort
executed the current policy with the biomedical tools (Methods) on a
fixed, stratified panel of 50 LIRICAL cases, producing an outcome,
reasoning trace, and tool record. Claude Opus 4.6 at xhigh reasoning
effort proposed a failure interpretation and candidate revision. A
clinical expert admitted a change only when it represented sound
clinical reasoning rather than a benchmark artifact and passed the
no-regression check. Accepted corrections entered the versioned policy;
when failure arose from returned evidence rather than its
interpretation, the tool specification was revised. The workflow, phase
criteria, and tools therefore co-evolved. Development stopped after
performance plateaued for at least 10 iterations. The public policy
required 42 iterations, of which 17 admitted configuration changes; only
the 50 LIRICAL development cases, and no PhenoPacket Store case,
informed the loop. The frozen artifact is the final policy.

\subsubsection{Benchmarks and clinical
cohort}\label{benchmarks-and-clinical-cohort}

liteOdyssey was evaluated on two public phenotype-first monogenic
rare-disease benchmarks and one private clinical cohort. Each public
case included clinical features and a known causal disease.

\textbf{LIRICAL} contains 370 cases across 252 diseases; approximately
45\% of its causal diseases are ultra-rare by Orphanet prevalence
mapping\citep{nguyen_diagnostic_2026}.

\textbf{PhenoPacket Store} comprised 873 cases: a stratified 300-case
sample (150 mapped and 150 unmapped), plus a 573-case extension covering
the store's broader disease distribution. \textbf{Mapped} indicates that
the causal disease was linked to Orphanet by our rarity-mapping
pipeline; \textbf{unmapped} indicates no link and serves as a proxy for
recently described, poorly indexed, or otherwise difficult-to-link
diseases. Across the corpus, 52.8\% of causal diseases were ultra-rare.

These benchmarks begin at clinical features, as a diagnostic workup
does, and span a broad ultra-rare disease load. Policy development used
only 50 LIRICAL cases and no PhenoPacket Store case.

\textbf{Undiagnosed Diseases Network (UDN) cohort.} The private cohort
included 515 patients referred after standard clinical workup did not
reach a diagnosis; each later received a UDN-established diagnosis. It
spanned 447 diagnoses across 15 US clinical sites (Appendix Tables A5
and A6). GPT-5.3 ran on an institutionally approved, secure Microsoft
Azure OpenAI instance. Fifty cases informed policy development and 465
were held out. The public-benchmark policy was extended through the same
PIHF process for atypical presentations and incomplete phenotype
information. A case-shape step routes atypical or multi-system
presentations to dedicated reasoning while typical cases retain the
public policy. Only aggregate results are reported; no patient-level
material is shared. The Vanderbilt University Medical Center
Institutional Review Board approved the study (\#172005).

The rarity-mapping pipeline, and per-benchmark prevalence distributions
are available at
https://github.com/minhha0510/Rare-Disease-Meta-Analysis-2026.

\subsubsection{Outcomes and scoring}\label{outcomes-and-scoring}

Primary metrics were disease \textbf{Recall@1} and \textbf{Recall@5}
under a standardized LLM-as-a-judge prompt adapted from
DeepRare\citep{zhao_agentic_2026}. For each case, the DeepRare-style
judge received the causal disease and aliases constructed by the
published expansion method\citep{zhao_agentic_2026}, plus the system's
ranked candidates. Recall@1 credits a match at rank one; Recall@5
credits a match within the top five.

Every case also received a deterministic \textbf{exact-OMIM cross-check}
comparing output identifiers with ground truth. Exact-OMIM is stricter
than the synonym-tolerant judge, so values are reported separately.
Before matching, non-canonical forms (bare digits, \texttt{MIM:}
prefixes, or a disease name in the identifier slot) were reconciled
against the gold standard uniformly across frozen model outputs. On
LIRICAL, the GPT-5.4 parametric baseline scored 35.1\% (130/370) under
the judge and 34.3\% (127/370) under exact-OMIM; liteOdyssey scored
58.6\% (217/370) and 55.9\% (207/370), respectively. All four ablation
configurations (Methods) use exact-OMIM.

DeepSeek-chat served as the DeepRare-style judge. Because models can
favor their own generations\citep{panickssery_llm_2024}, no judge shared
a model family with the diagnostic system. DeepSeek-family runs were
therefore re-scored with the same prompt by Claude Sonnet and are
reported using exact-OMIM and the Sonnet judge.

Physicians blinded to model identity additionally reviewed the UDN
outputs. Two physicians independently scored a prespecified 120-case
sample, one liteOdyssey and one parametric-baseline output per case (240
randomized outputs), on the Bond differential-diagnosis scale (0, no
close suggestion; 2, related but unlikely helpful; 3, closely related;
4, very close; 5, exact diagnosis)\citep{bond_differential_2012}. A
third blinded physician adjudicated 66 output-level disagreements
spanning 51 cases by selecting a prior score or providing an independent
score. Prespecified sampling strata combined strict outcomes with judge
confidence; inverse selection probabilities weighted scores to the
515-case cohort. The primary comparison was the weighted mean paired
Bond-score difference under a design-based Wald test. Four adjudication
scenarios served as sensitivity analyses, and quadratic-weighted κ
summarized inter-rater agreement.

Full-system and matched-baseline Recall@1 and Recall@5 were compared by
exact McNemar tests on case-level pairs, with 95\% confidence intervals
for proportion differences from a 20,000-iteration bootstrap. Weighted
physician-review proportions use pointwise 95\% Wilson intervals;
gene-level UDN comparisons are Bonferroni-corrected across both recall
metrics. All eligible cases produced scorable outcomes and were retained
in the outcome analyses.

\subsubsection{Ablation, cross-model evaluation, and
reproducibility}\label{ablation-cross-model-evaluation-and-reproducibility}

Two contrastive evaluations tested policy attribution and model
transfer.

\textbf{Ablation ladder (progressively stronger configurations).} Four
cumulative configurations tested whether the workflow adds value beyond
optimized prompting and access to the same information. Each used
GPT-5.4 and deterministic exact-OMIM scoring (Methods) on 370 LIRICAL,
150 PhenoPacket Store unmapped, and 573 extension cases with frozen
prompts and scoring: (i) the tool-free parametric baseline, prompted
from clinical features using a strategy adapted from prior
work\citep{zhao_agentic_2026} and answering from internal knowledge;
(ii) a hardened tool-free prompt generated without human authoring by
GPT-5.5 at xhigh reasoning effort through automatic generation,
self-critique, revision, freezing, and adversarial review; (iii) that
frozen prompt given the source families and endpoint roots selected
during policy development, but without liteOdyssey's policy, query
discipline, or adjudication; and (iv) the full policy and engineered
tools. Holding the execution model fixed identifies the added value of
prompting, source access, and the policy-tool system. Because the
configurations are cumulative, increments are interpreted as a monotonic
ladder rather than separable component effects (Results; Appendix Table
A2). Appendix C provides the meta-prompts and source-access
specification.

\textbf{Cross-model evaluation.} The same policy and tools were run with
Qwen3.6-35B-A3B, DeepSeek-V4-Flash-Preview, DeepSeek-V4-Pro-Preview,
Claude Opus 4.6 (LIRICAL, PU and PM only), and GPT-5.4 on shared LIRICAL
and PhenoPacket Store cases (mapped, unmapped, and extension where
available). The model was the only varying component; policy, tools,
workflow, prompt, and scoring were fixed. Recorded outcomes were
Recall@1, judged-R@1 verdict agreement, and per-cohort tool calls.

Public-benchmark runs used GPT-5.4, open-weight Qwen3.6-35B-A3B, and
DeepSeek-V4 Flash-Preview and Pro-Preview at high reasoning effort. The
release will record model, benchmark, rarity-mapping, tool-library,
judge, and prompt versions; the public policy will be released with
publication to encourage reproducibility.

\textbf{Deployment footprint.} Footprint includes bytes on disk for the
policy, tool runtime, and phenotype-to-HPO mapping needed for public
benchmarks. Users may cache optional OMIM and HPO tables, acquired under
their own OMIM licence. On this boundary, liteOdyssey occupies 46 MB,
including 2.5 MB redistributed with the package. PIHF performed no
weight-space training; compute consisted of inference-time model calls
and deterministic tool execution, and provider hardware was not exposed
through the model APIs.

\textbf{Website}: https://liteodyssey.org/

\subsection{Acknowledgements.}\label{acknowledgements.}

The authors are grateful to the patients for participating in the
Undiagnosed Diseases Network.

\subsection{Funding.}\label{funding.}

This work was supported in part by the National Institutes of Health
Common Fund, grant 15-HG-0130 from the National Human Genome Research
Institute, U01NS134349 from the National Institute of Neurological
Disorders and Stroke, R00LM014429 and K99LM014428 from the National
Library of Medicine, R35GM162668 from the National Institute of General
Medical Sciences, and the Potocsnak Center for Undiagnosed and Rare
Disorders. The NIH Undiagnosed Diseases Program is supported by the
Intramural Research Program of the National Human Genome Research
Institute. The content is solely the responsibility of the authors and
does not necessarily represent the official views of the National
Institutes of Health. The funders had no role in the study design, data
collection and analysis, decision to publish, or preparation of the
manuscript.

\subsection{Author contributions.}\label{author-contributions.}

C.S. and T.A.C conceived the study. M.H., C.S., and T.A.C. designed the
experiments. M.H. performed the experiments. M.H., E.T.G., B.A.S.,
T.A.C., and C.S. collected and analyzed the data. M.H. developed the
computational methods and performed the statistical analyses. M.H.,
E.T.G., B.A.S., K.W.B., C.T.Y., F.M., H.X., W.C.S., C.Y., W.W., A.W.,
L.B., J.F.P., L.L., S.M., R.H., T.A.C., and C.S. interpreted the
results. M.H. prepared the figures. M.H. prepared the manuscript with
input from all other authors. R.H., T.A.C., and C.S. supervised the
study and acquired funding. All authors discussed the results and
reviewed and approved the manuscript.

\subsection{Competing interests.}\label{competing-interests.}

The authors declare no competing interests.

\subsection{Data availability.}\label{data-availability.}

The benchmark case lists, development and held-out splits, model
outputs, and scoring artifacts used in this study will be released upon
peer-reviewed publication. LIRICAL and PhenoPacket Store cases are
publicly available from their original sources, and the rarity-mapping
pipeline is available at
https://github.com/minhha0510/Rare-Disease-Meta-Analysis-2026.

The Undiagnosed Diseases Network data used in this study contain
sensitive patient information. De-identified patient data, including
phenotypic and genomic data, are deposited in the
\href{https://www.ncbi.nlm.nih.gov/gap}{database of Genotypes and
Phenotypes (dbGaP)} maintained by the National Institutes of Health. To
explore data available in the latest release, visit the
\href{https://www.ncbi.nlm.nih.gov/projects/gap/cgi-bin/study.cgi?study_id=phs001232.v8.p3}{UDN
study page in dbGaP}. Individuals interested in accessing UDN data
through dbGaP should submit a data access request. Detailed instructions
for this process can be found on the NIH Scientific Data Sharing
website:
\href{https://sharing.nih.gov/accessing-data/accessing-genomic-data/how-to-request-and-access-datasets-from-dbgap}{How
to Request and Access Datasets from dbGaP}.

\subsection{Code availability.}\label{code-availability.}

The PIHF-derived public-benchmark policy and the software needed to
execute the policy and reproduce the reported analyses will be released
open-source upon peer-reviewed publication. The released version,
repository address, licence and persistent identifier will be added to
this section before publication. The rarity-mapping pipeline used in
this study is available at
https://github.com/minhha0510/Rare-Disease-Meta-Analysis-2026.

\bibliography{refs}

\clearpage

\subsection{Appendix A. Auditable reasoning and worked
cases}\label{appendix-a.-auditable-reasoning-and-worked-cases}

liteOdyssey returns a ranked differential diagnosis together with the
reasoning process and tool traces that produced it. For each case, the
reasoning process lists the clinical features considered, genes and
diseases evaluated, biomedical resources queried, evidence supporting or
weakening each candidate hypothesis, and rationale for the final
ranking. This process provides a clinical audit trail showing what
evidence was gathered and how it influenced the differential.

For example, in one LIRICAL case involving Myhre syndrome, the
parametric baseline matched the patient's skeletal and sclerosing
features to unrelated skeletal dysplasias and did not rank the correct
diagnosis highly. In comparison, liteOdyssey connected the patient's
phenotypes to the gene SMAD4, identified Myhre syndrome through disease
lookup, checked the strength of the gene--disease relationship, and kept
the more specific diagnosis as the top-ranked candidate after reflective
review, reporting high confidence. On the other hand, in a LIRICAL case
of Nijmegen breakage syndrome, the parametric baseline named the
diagnosis directly and reported high confidence. liteOdyssey identified
the disease class correctly but adjudicated to NHEJ1-related Cernunnos
deficiency, the close phenocopy of the true disease, and ranked it first
at moderate confidence. The trace records the literature retrieval that
raised the alternative and the adjudication that preferred it, and the
lowered confidence rating marks the case as unresolved.

Across the 1,243 public benchmark cases, regression over the
off-the-shelf baseline is rare (1.4\% - 2.4\%) and the losses also
announce themselves: 82\% of regression cases (70/85) carry
low-to-moderate confidence, and only 15 regressions occurred at high
confidence. Therefore, liteOdyssey provides differential diagnoses that
are substantially more often correct, together with a marker on the
cases where they are not.

The worked cases below are public LIRICAL cases in which the full system
and its own parametric baseline (the same GPT-5.4 model, differing only
in the diagnostic policy and its tools) disagreed. De-identified:
disease name + OMIM only; no case identifiers, iteration names, or
internal tool labels. Myhre syndrome and Nijmegen breakage syndrome are
summarized above. Myhre syndrome is traced in full below as a worked
example along five dimensions a clinician would inspect: the phenotypes
that seeded the hypothesis, the candidates considered, the evidence that
supported or weakened each, where uncertainty remained, and what a
reviewer could inspect. Together, these dimensions show concretely how
the policy and tools reach a diagnosis the unaided model misses. The
remaining cases are documented more briefly.

\subsubsection{Rescues: where the policy and tools earn the
diagnosis}\label{rescues-where-the-policy-and-tools-earn-the-diagnosis}

\textbf{Myhre syndrome (OMIM 139210; causal gene SMAD4).}
\emph{(summarized above.)} \emph{Phenotypes that seeded the hypothesis:}
a sclerosing, hyperostotic skeletal presentation with a distinctive
signature, enlarged vertebral pedicles and a thickened calvaria,
alongside brachydactyly, joint stiffness, and conductive hearing loss.
\emph{Candidates considered:} the parametric baseline pattern-matched
this presentation to a basket of hyperostotic and filamin-related
dysplasias, ranking Lenz--Majewski hyperostotic dwarfism first, with
frontometaphyseal dysplasia, otopalatodigital-spectrum disorders, and
Melnick--Needles syndrome below it, and never reached the correct
disease. The full system instead surfaced SMAD4 and adjudicated it
against its nearest competitor, NFIX (Marshall--Smith syndrome).
\emph{Evidence that supported or weakened each candidate:} the full
system's phenotype-to-gene retrieval quantified SMAD4 as the top-scoring
gene (specificity score 1.1137 across 11 matched terms, summing inverse
gene-annotation counts); an independent disease-level lookup returned
Myhre syndrome as the top hit; and gene--disease validity and
population-constraint evidence confirmed SMAD4 as a Definitive,
constraint-intolerant dominant gene. In adjudication, only Myhre
syndrome accounted for the thickened calvaria, enlarged vertebral
pedicles, brachydactyly, and joint stiffness, which NFIX-related disease
does not. \emph{Where uncertainty remained:} little: the system reported
HIGH confidence, the specific skeletal signature anchoring SMAD4 where
the unaided model had defaulted to the more familiar hyperostotic
disorders. \emph{What a reviewer could inspect:} a quantified evidence
chain assembled across 29 tool calls (phenotype-to-gene ranking, disease
lookup, gene--disease validity, population constraint), each step
traceable to its public source; Myhre syndrome was ranked first.

\textbf{Cornelia de Lange syndrome type 3 (OMIM 610759).} The baseline
correctly recognized the Cornelia de Lange syndrome family but defaulted
to the canonical gene of a different subtype (NIPBL, type 1), which the
diagnosis-level judge scores as a miss. The full system's trace shows a
focused re-query on the discriminating distal-limb and periocular
features, curly eyelashes, proximal thumbs, and short fourth and fifth
metacarpals, holding the correct subtype gene above the canonical one;
it ranked Cornelia de Lange syndrome type 3 first. Retrieved,
case-specific evidence overrode a strong but wrong textbook default.

\subsection{Appendix B. Diagnostic
policy}\label{appendix-b.-diagnostic-policy}

\subsubsection{B.1 The eight-phase reasoning
workflow}\label{b.1-the-eight-phase-reasoning-workflow}

The policy produced by PIHF (Methods, `Policy Iteration with Human
Feedback') is an eight-phase clinical reasoning workflow, executed
end-to-end by a single reasoning-capable model. The workflow is
structured but not strictly linear: the system can revisit earlier
phases when new evidence changes the working differential or when
uncertainty remains high. Each phase specifies which biomedical tools
(Methods) it draws on, so the phases below also indicate how
tool-derived evidence enters the diagnostic process.

\textbf{Phase 0: Pattern recognition.} The system reviews the HPO term
list to form an initial hypothesis about the likely disease family,
anchored on the most specific features.

\textbf{Phase 1: Phenotype-based candidate gene generation.} The system
queries the phenotype-to-gene tool, with rare and specific terms
weighted more heavily than common terms: each term's weight is the
inverse of its number of annotated genes (denoted IC score in the policy
text; Appendix Figure B2). The output is an initial gene shortlist based
on similarity between the patient's clinical features and known
gene-phenotype associations.

\textbf{Phase 2: Evidence triage.} Each gene on the shortlist is
evaluated for curated gene-disease validity. Genes with definitive,
strong, or moderate evidence are prioritized for deeper evaluation;
genes with limited or disputed evidence are deprioritized unless they
remain uniquely consistent with the phenotype.

\textbf{Phase 3: Deep investigation.} For prioritized gene candidates,
the system retrieves disease-level evidence on inheritance, age of
onset, associated clinical features, and related syndromes, then builds
a case for or against each candidate by comparing the disease's expected
presentation against the patient's observed features.

\textbf{Phase 4: Confidence assessment.} The system assesses whether the
top candidate explains the full phenotype, whether any feature cluster
remains unexplained, and whether a related syndrome could provide a
better fit.

\textbf{Phase 5: Corrective search.} When confidence is low or a
clinical feature cluster remains unexplained, the system performs
targeted literature and disease-family searches for recently described
or poorly indexed candidates. This step addresses cases in which the
initial ranking missed the correct answer because the disease was too
new or too rare to appear prominently.

\textbf{Phase 6: Reflective adjudication.} Before finalizing the
differential, the system re-examines the leading candidate against its
competitors, including those with greater phenotypic specificity or less
common but more clinically consistent presentations. This step targets a
recurrent error mode: when two related candidates are plausible, a model
may otherwise favor the more common or better-annotated disorder over a
rarer but more phenotypically specific alternative. The system asks
three pre-defined questions: whether the leading diagnosis explains all
major feature clusters, whether a competing candidate provides a more
specific phenotypic fit, and whether targeted literature or
disease-family search could alter the ranking.

\textbf{Phase 7: Final output.} The system produces a ranked top-five
differential diagnosis. For each candidate, it reports the disease,
associated gene, supporting evidence, confidence classification, and the
rationale for its position in the ranking.

Because the model executes these phases in sequence, the workflow yields
a single continuous reasoning process: every tool call, intermediate
ranking, and reflective re-check is recorded in order. This record
provides the basis for clinical auditability (Appendix A): a reviewing
clinician can follow, step by step, how the differential was constructed
and challenge any inference that lacks retrieved support.

\subsubsection{B.2 The clinical reasoning component of the
policy}\label{b.2-the-clinical-reasoning-component-of-the-policy}

In this section, we present the natural-language component of the
policy, comprising the clinical reasoning patterns. The policy used to
evaluate the public benchmarks will be open-sourced, along with the
datasets, upon publication of this paper to encourage reproducibility.
The complete policy, as released, is reproduced in Appendix Figure B2.

\subsection{Appendix C. Hardened baseline and source
access}\label{appendix-c.-hardened-baseline-and-source-access}

Ablation condition (ii) used a hardened diagnostic prompt generated
autonomously, with no human authoring, by a frontier model at high
reasoning effort through a generate, self-critique, revise, freeze
procedure with adversarial review by the same model. The three
meta-prompts that drove that procedure, the frozen prompt they produced,
and the source-access specification added for condition (iii) are
reproduced so the strength of the prompt-only baseline can be judged
directly.

\clearpage
\floatplacement{figure}{H}

\subsection{Supplementary Information:
liteOdyssey}\label{supplementary-information-liteodyssey}

\begin{figure}
\centering
\includegraphics[width=0.85\linewidth,height=\textheight,keepaspectratio,alt={Appendix Figure A1. Full liteOdyssey system Recall@1 and Recall@5 under the same policy and tools on GPT-5.4 versus Claude Opus 4.6 across 670 cases (370 LIRICAL plus 300 PhenoPacket Store, mapped and unmapped). Scored with the same LLM-as-a-judge protocol as Figure 2.}]{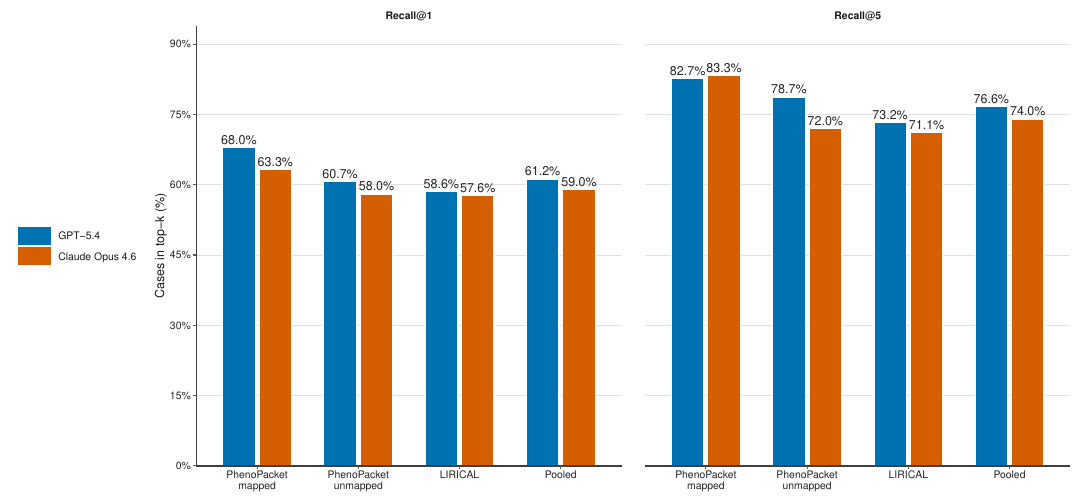}
\caption{Appendix Figure A1. Full liteOdyssey system Recall@1 and
Recall@5 under the same policy and tools on GPT-5.4 versus Claude Opus
4.6 across 670 cases (370 LIRICAL plus 300 PhenoPacket Store, mapped and
unmapped). Scored with the same LLM-as-a-judge protocol as Figure 2.}
\end{figure}

\begin{figure}
\centering
\includegraphics[width=0.85\linewidth,height=\textheight,keepaspectratio,alt={Appendix Figure A2. Gene-level prediction: causal-gene ranking, full system versus parametric baseline. (a) On the full PhenoPacket Store (871 cases with causal-gene truth; GPT-5.4), the full system ranked the causal gene first in 61.2\% of cases and within the top five in 77.8\%, versus 19.2\% and 28.4\% for the parametric baseline (+42.0 and +49.5 points); the open-weights Qwen3.6 showed the same pattern (60.0\% versus 8.4\% at Recall@1). (b) On the 515-case UDN cohort (GPT-5.3), the gain was attenuated and concentrated at Recall@5 (33.8\% versus 28.5\%; paired McNemar p = 0.0049 after Bonferroni correction), while the Recall@1 difference was not significant (21.9\% versus 19.8\%; p = 0.21). Gene-level accuracy is scored only where causal-gene ground truth exists (the PhenoPacket Store and UDN; LIRICAL is defined at the OMIM-disease level and does not include gene labels). Two PhenoPacket Store extension cases (PE\_015, PE\_016) carry no causal-gene truth and are excluded from both arms, so the extension contributes 571 of its 573 cases.}]{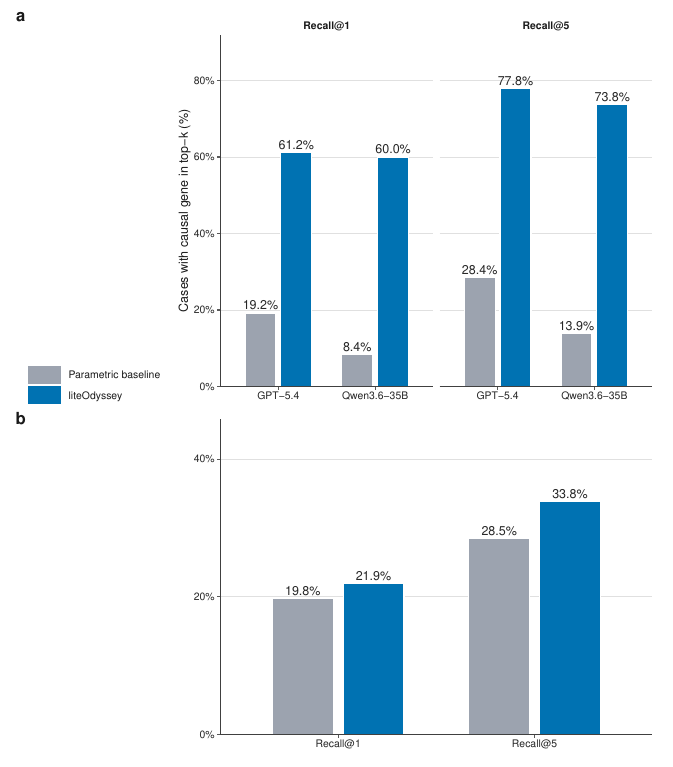}
\caption{Appendix Figure A2. Gene-level prediction: causal-gene ranking,
full system versus parametric baseline. (a) On the full PhenoPacket
Store (871 cases with causal-gene truth; GPT-5.4), the full system
ranked the causal gene first in 61.2\% of cases and within the top five
in 77.8\%, versus 19.2\% and 28.4\% for the parametric baseline (+42.0
and +49.5 points); the open-weights Qwen3.6 showed the same pattern
(60.0\% versus 8.4\% at Recall@1). (b) On the 515-case UDN cohort
(GPT-5.3), the gain was attenuated and concentrated at Recall@5 (33.8\%
versus 28.5\%; paired McNemar p = 0.0049 after Bonferroni correction),
while the Recall@1 difference was not significant (21.9\% versus 19.8\%;
p = 0.21). Gene-level accuracy is scored only where causal-gene ground
truth exists (the PhenoPacket Store and UDN; LIRICAL is defined at the
OMIM-disease level and does not include gene labels). Two PhenoPacket
Store extension cases (PE\_015, PE\_016) carry no causal-gene truth and
are excluded from both arms, so the extension contributes 571 of its 573
cases.}
\end{figure}

\textbf{Appendix Table A1. Disease distribution across development and
held-out evaluation cases.}

{\def\LTcaptype{none} 
\begin{longtable}[]{@{}
  >{\raggedright\arraybackslash}p{(\linewidth - 8\tabcolsep) * \real{0.2000}}
  >{\raggedright\arraybackslash}p{(\linewidth - 8\tabcolsep) * \real{0.2000}}
  >{\raggedright\arraybackslash}p{(\linewidth - 8\tabcolsep) * \real{0.2000}}
  >{\raggedright\arraybackslash}p{(\linewidth - 8\tabcolsep) * \real{0.2000}}
  >{\raggedright\arraybackslash}p{(\linewidth - 8\tabcolsep) * \real{0.2000}}@{}}
\toprule\noalign{}
\begin{minipage}[b]{\linewidth}\raggedright
Cohort
\end{minipage} & \begin{minipage}[b]{\linewidth}\raggedright
Cases
\end{minipage} & \begin{minipage}[b]{\linewidth}\raggedright
Unique diseases
\end{minipage} & \begin{minipage}[b]{\linewidth}\raggedright
Development
\end{minipage} & \begin{minipage}[b]{\linewidth}\raggedright
Held out
\end{minipage} \\
\midrule\noalign{}
\endhead
\bottomrule\noalign{}
\endlastfoot
LIRICAL & 370 & 252 & 50 cases (43 diseases) & 320 cases \\
PhenoPacket Store (mapped) & 150 & 125 & 0 & 150 \\
PhenoPacket Store (unmapped) & 150 & 139 & 0 & 150 \\
PhenoPacket Store (extension) & 573 & 297 & 0 & 573 \\
Public total (deduplicated union)1 & 1,243 & 722 & 50 cases (43
diseases) & 1,193 cases \\
UDN (private) & 515 & 447 & 50 cases (50 diseases) & 465 cases (407
diseases) \\
\end{longtable}
}

\textbf{Appendix Table A2. Progressively stronger configurations on
LIRICAL (370 cases), PhenoPacket Store unmapped (150 cases) and
PhenoPacket Store extension (573 cases) (exact-OMIM cross-check), all
executed on the same model (GPT-5.4). Each configuration adds a layer to
the one above; the configurations are cumulative and their individual
contributions are not separable.}

{\def\LTcaptype{none} 
\begin{longtable}[]{@{}
  >{\raggedright\arraybackslash}p{(\linewidth - 6\tabcolsep) * \real{0.2500}}
  >{\raggedright\arraybackslash}p{(\linewidth - 6\tabcolsep) * \real{0.2500}}
  >{\raggedright\arraybackslash}p{(\linewidth - 6\tabcolsep) * \real{0.2500}}
  >{\raggedright\arraybackslash}p{(\linewidth - 6\tabcolsep) * \real{0.2500}}@{}}
\toprule\noalign{}
\begin{minipage}[b]{\linewidth}\raggedright
Cohort
\end{minipage} & \begin{minipage}[b]{\linewidth}\raggedright
Configuration
\end{minipage} & \begin{minipage}[b]{\linewidth}\raggedright
Recall@1
\end{minipage} & \begin{minipage}[b]{\linewidth}\raggedright
Recall@5
\end{minipage} \\
\midrule\noalign{}
\endhead
\bottomrule\noalign{}
\endlastfoot
LIRICAL & Base model (clinical features only) & 34.3\% & 48.6\% \\
LIRICAL & + hardened, autonomously generated diagnostic prompt (no
tools) & 36.8\% & 52.7\% \\
LIRICAL & + access to the same PIHF-optimized public biomedical sources
& 44.9\% & 61.9\% \\
LIRICAL & Full liteOdyssey policy and engineered tools & 55.9\% &
72.7\% \\
PhenoPacket Store unmapped & Base model (clinical features only) & 6.0\%
& 8.7\% \\
PhenoPacket Store unmapped & + hardened, autonomously generated
diagnostic prompt (no tools) & 4.7\% & 9.3\% \\
PhenoPacket Store unmapped & + access to the same PIHF-optimized public
biomedical sources & 30.7\% & 44.0\% \\
PhenoPacket Store unmapped & Full liteOdyssey policy and engineered
tools & 60.7\% & 78.7\% \\
PhenoPacket Store extension & Base model (clinical features only) &
10.1\% & 13.8\% \\
PhenoPacket Store extension & + hardened, autonomously generated
diagnostic prompt (no tools) & 10.1\% & 14.5\% \\
PhenoPacket Store extension & + access to the same PIHF-optimized public
biomedical sources & 48.7\% & 60.4\% \\
PhenoPacket Store extension & Full liteOdyssey policy and engineered
tools & 56.4\% & 74.7\% \\
\end{longtable}
}

\textbf{Appendix Table A3. Judge-style versus exact-OMIM scoring on the
development-excluded rows (the 320 held-out LIRICAL and 465 held-out UDN
cases).}

{\def\LTcaptype{none} 
\begin{longtable}[]{@{}
  >{\raggedright\arraybackslash}p{(\linewidth - 12\tabcolsep) * \real{0.1429}}
  >{\raggedright\arraybackslash}p{(\linewidth - 12\tabcolsep) * \real{0.1429}}
  >{\raggedright\arraybackslash}p{(\linewidth - 12\tabcolsep) * \real{0.1429}}
  >{\raggedright\arraybackslash}p{(\linewidth - 12\tabcolsep) * \real{0.1429}}
  >{\raggedright\arraybackslash}p{(\linewidth - 12\tabcolsep) * \real{0.1429}}
  >{\raggedright\arraybackslash}p{(\linewidth - 12\tabcolsep) * \real{0.1429}}
  >{\raggedright\arraybackslash}p{(\linewidth - 12\tabcolsep) * \real{0.1429}}@{}}
\toprule\noalign{}
\begin{minipage}[b]{\linewidth}\raggedright
Cohort
\end{minipage} & \begin{minipage}[b]{\linewidth}\raggedright
Endpoint
\end{minipage} & \begin{minipage}[b]{\linewidth}\raggedright
Workflow
\end{minipage} & \begin{minipage}[b]{\linewidth}\raggedright
N
\end{minipage} & \begin{minipage}[b]{\linewidth}\raggedright
Judge
\end{minipage} & \begin{minipage}[b]{\linewidth}\raggedright
Exact OMIM
\end{minipage} & \begin{minipage}[b]{\linewidth}\raggedright
Judge - exact, pp
\end{minipage} \\
\midrule\noalign{}
\endhead
\bottomrule\noalign{}
\endlastfoot
LIRICAL oos320 & R@1 & Full system & 320 & 184/320 (57.5\%) & 174/320
(54.4\%) & +3.1 \\
LIRICAL oos320 & R@1 & Parametric & 320 & 109/320 (34.1\%) & 106/320
(33.1\%) & +1.0 \\
LIRICAL oos320 & R@5 & Full system & 320 & 230/320 (71.9\%) & 228/320
(71.2\%) & +0.7 \\
LIRICAL oos320 & R@5 & Parametric & 320 & 163/320 (50.9\%) & 153/320
(47.8\%) & +3.1 \\
UDN oos465 & R@1 & Full system & 465 & 89/465 (19.1\%) & 89/465 (19.1\%)
& +0.0 \\
UDN oos465 & R@1 & Parametric & 465 & 72/465 (15.5\%) & 66/465 (14.2\%)
& +1.3 \\
UDN oos465 & R@5 & Full system & 465 & 139/465 (29.9\%) & 139/465
(29.9\%) & +0.0 \\
UDN oos465 & R@5 & Parametric & 465 & 110/465 (23.7\%) & 101/465
(21.7\%) & +2.0 \\
\end{longtable}
}

\textbf{Appendix Table A4. liteOdyssey biomedical public tool library:
the clinical question each tool answers, its source, and the evidence it
returns (the five phenotype-first knowledge tools used on the public
benchmarks; the two variant-mode components are described in Methods,
`Biomedical tool library').}

{\def\LTcaptype{none} 
\begin{longtable}[]{@{}
  >{\raggedright\arraybackslash}p{(\linewidth - 6\tabcolsep) * \real{0.2500}}
  >{\raggedright\arraybackslash}p{(\linewidth - 6\tabcolsep) * \real{0.2500}}
  >{\raggedright\arraybackslash}p{(\linewidth - 6\tabcolsep) * \real{0.2500}}
  >{\raggedright\arraybackslash}p{(\linewidth - 6\tabcolsep) * \real{0.2500}}@{}}
\toprule\noalign{}
\begin{minipage}[b]{\linewidth}\raggedright
Tool
\end{minipage} & \begin{minipage}[b]{\linewidth}\raggedright
Clinical question
\end{minipage} & \begin{minipage}[b]{\linewidth}\raggedright
Source
\end{minipage} & \begin{minipage}[b]{\linewidth}\raggedright
Evidence returned
\end{minipage} \\
\midrule\noalign{}
\endhead
\bottomrule\noalign{}
\endlastfoot
dx-monarch & Which genes match the patient's phenotypes? & Monarch
Initiative & Specificity-weighted candidate gene list (inverse
gene-annotation count per term) \\
dx-omim & What diseases does this gene cause / which genes are in this
disease family? & OMIM (local files: genemap2, morbidmap, mimTitles; HPO
annotations phenotype.hpoa) & Disease records (inheritance, age of
onset), keyword search, and disease-series enumeration \\
dx-gene-evidence & Is the gene intolerant of loss-of-function / missense
variation? & gnomAD (pLI, LOEUF, missense\_z) + OMIM & Per-gene
constraint metrics + OMIM diseases \\
dx-clingen & How strong is the curated gene-disease link / dosage
sensitivity? & ClinGen (local reference CSVs) & Gene-disease validity +
haploinsufficiency classification \\
dx-pubmed & Are there relevant case reports for this presentation? &
PubMed (NCBI E-utilities) & Relevant abstracts \\
\end{longtable}
}

\textbf{Appendix Table A5. UDN cohort composition.}

{\def\LTcaptype{none} 
\begin{longtable}[]{@{}
  >{\raggedright\arraybackslash}p{(\linewidth - 2\tabcolsep) * \real{0.5000}}
  >{\raggedright\arraybackslash}p{(\linewidth - 2\tabcolsep) * \real{0.5000}}@{}}
\toprule\noalign{}
\begin{minipage}[b]{\linewidth}\raggedright
Property
\end{minipage} & \begin{minipage}[b]{\linewidth}\raggedright
Value
\end{minipage} \\
\midrule\noalign{}
\endhead
\bottomrule\noalign{}
\endlastfoot
Patients & 515 (50 development, 465 held out) \\
Distinct diagnoses & 447 (407 among the held-out patients) \\
Diagnoses seen in a single patient & 411 of 447 \\
Most frequent diagnosis & ReNU syndrome, 25 patients \\
Clinical sites & 15 geographically distributed US sites (Appendix Table
A6) \\
\end{longtable}
}

\textbf{Appendix Table A6. UDN geographic distribution across 15 US
clinical sites, shown for all 515 patients, the 50 development patients,
and the 465 held-out patients (the four smallest sites, 1--2 patients
each, are grouped). The 50 development patients span 8 of the 15 sites.}

{\def\LTcaptype{none} 
\begin{longtable}[]{@{}
  >{\raggedright\arraybackslash}p{(\linewidth - 6\tabcolsep) * \real{0.2500}}
  >{\raggedright\arraybackslash}p{(\linewidth - 6\tabcolsep) * \real{0.2500}}
  >{\raggedright\arraybackslash}p{(\linewidth - 6\tabcolsep) * \real{0.2500}}
  >{\raggedright\arraybackslash}p{(\linewidth - 6\tabcolsep) * \real{0.2500}}@{}}
\toprule\noalign{}
\begin{minipage}[b]{\linewidth}\raggedright
Clinical site
\end{minipage} & \begin{minipage}[b]{\linewidth}\raggedright
All 515
\end{minipage} & \begin{minipage}[b]{\linewidth}\raggedright
Dev 50
\end{minipage} & \begin{minipage}[b]{\linewidth}\raggedright
Held-out 465
\end{minipage} \\
\midrule\noalign{}
\endhead
\bottomrule\noalign{}
\endlastfoot
NIH Undiagnosed Diseases Program & 110 (21.4\%) & 16 (32.0\%) & 94
(20.2\%) \\
Vanderbilt University Medical Center & 70 (13.6\%) & 10 (20.0\%) & 60
(12.9\%) \\
Duke University Medical Center & 67 (13.0\%) & 3 (6.0\%) & 64
(13.8\%) \\
UDN Central & 64 (12.4\%) & 6 (12.0\%) & 58 (12.5\%) \\
Baylor College of Medicine & 60 (11.7\%) & 4 (8.0\%) & 56 (12.0\%) \\
Stanford University & 46 (8.9\%) & 6 (12.0\%) & 40 (8.6\%) \\
University of Washington and Seattle Children's Hospital & 29 (5.6\%) &
0 (0\%) & 29 (6.2\%) \\
University of California, Los Angeles & 20 (3.9\%) & 3 (6.0\%) & 17
(3.7\%) \\
Harvard-affiliated Hospitals (BCH, BWH, MGH) & 19 (3.7\%) & 2 (4.0\%) &
17 (3.7\%) \\
Washington University in St.~Louis & 18 (3.5\%) & 0 (0\%) & 18
(3.9\%) \\
University of Miami & 7 (1.4\%) & 0 (0\%) & 7 (1.5\%) \\
Four additional sites (1--2 patients each) & 5 (1.0\%) & 0 (0\%) & 5
(1.1\%) \\
\end{longtable}
}

\subsubsection{Appendix Figure B2. The clinical reasoning component of
the
policy}\label{appendix-figure-b2.-the-clinical-reasoning-component-of-the-policy}

Appendix Figure B2. The clinical reasoning component of the policy. The
natural-language policy that guides liteOdyssey's diagnostic reasoning,
shown in full as released with the public package. The policy instructs
the model to reason as a clinical geneticist: it states the scope for
cases with and without genetic information, ranks candidates in two
tiers (tool-based evidence leads between gene families; clinical
phenotype-fit judgment leads within a family), admits annotation-gap
rescue candidates only when they win on evidence at the appropriate
tier, and executes the eight-phase workflow summarized in Appendix B.1.
The same policy, unchanged, produced the public-benchmark and
development-excluded results in Figures 2 and 3.

\includegraphics[width=1\linewidth,height=\textheight,keepaspectratio]{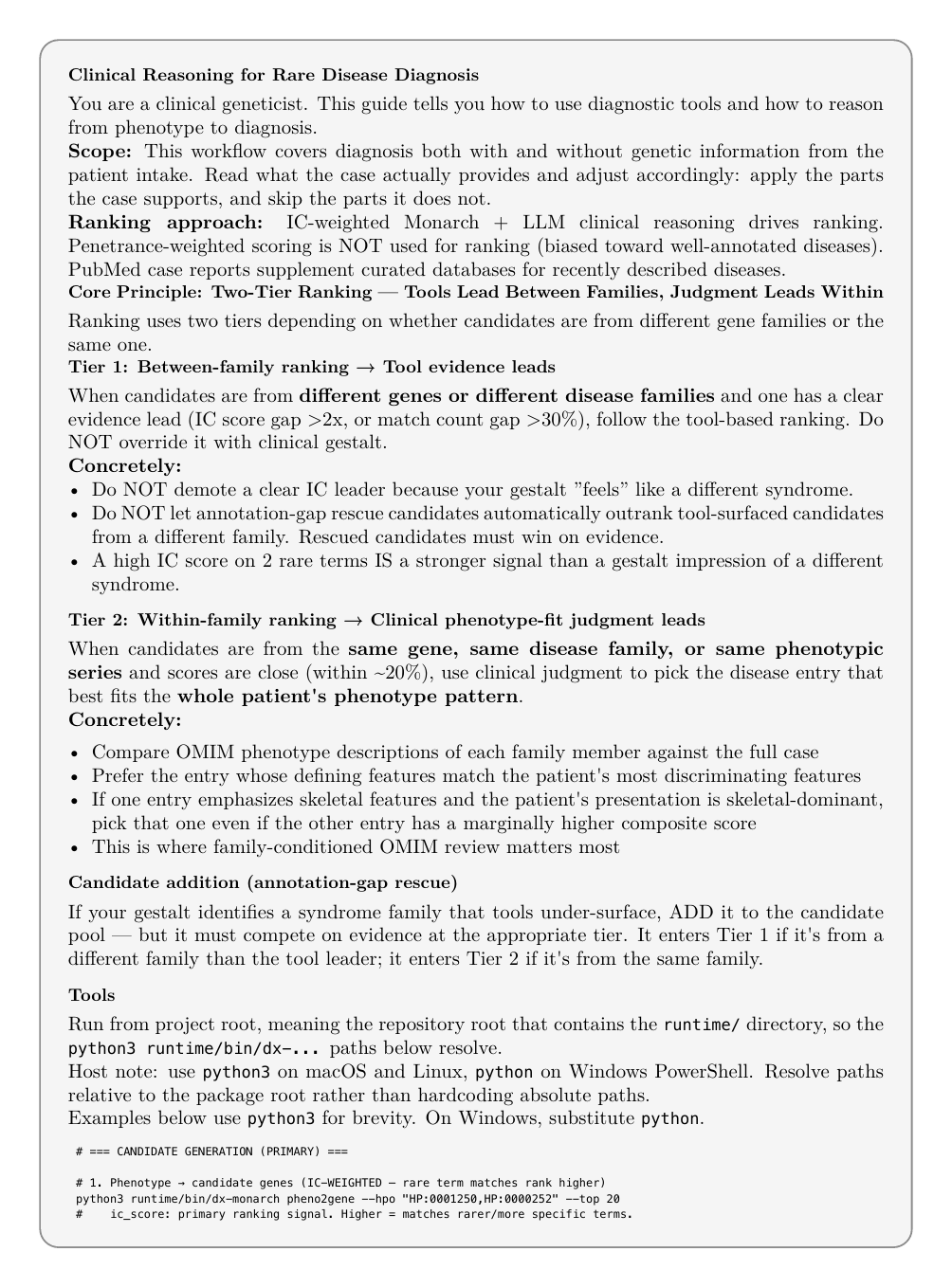}

Appendix Figure B2 (continued, page 2 of 12).

\includegraphics[width=0.93\linewidth,height=\textheight,keepaspectratio]{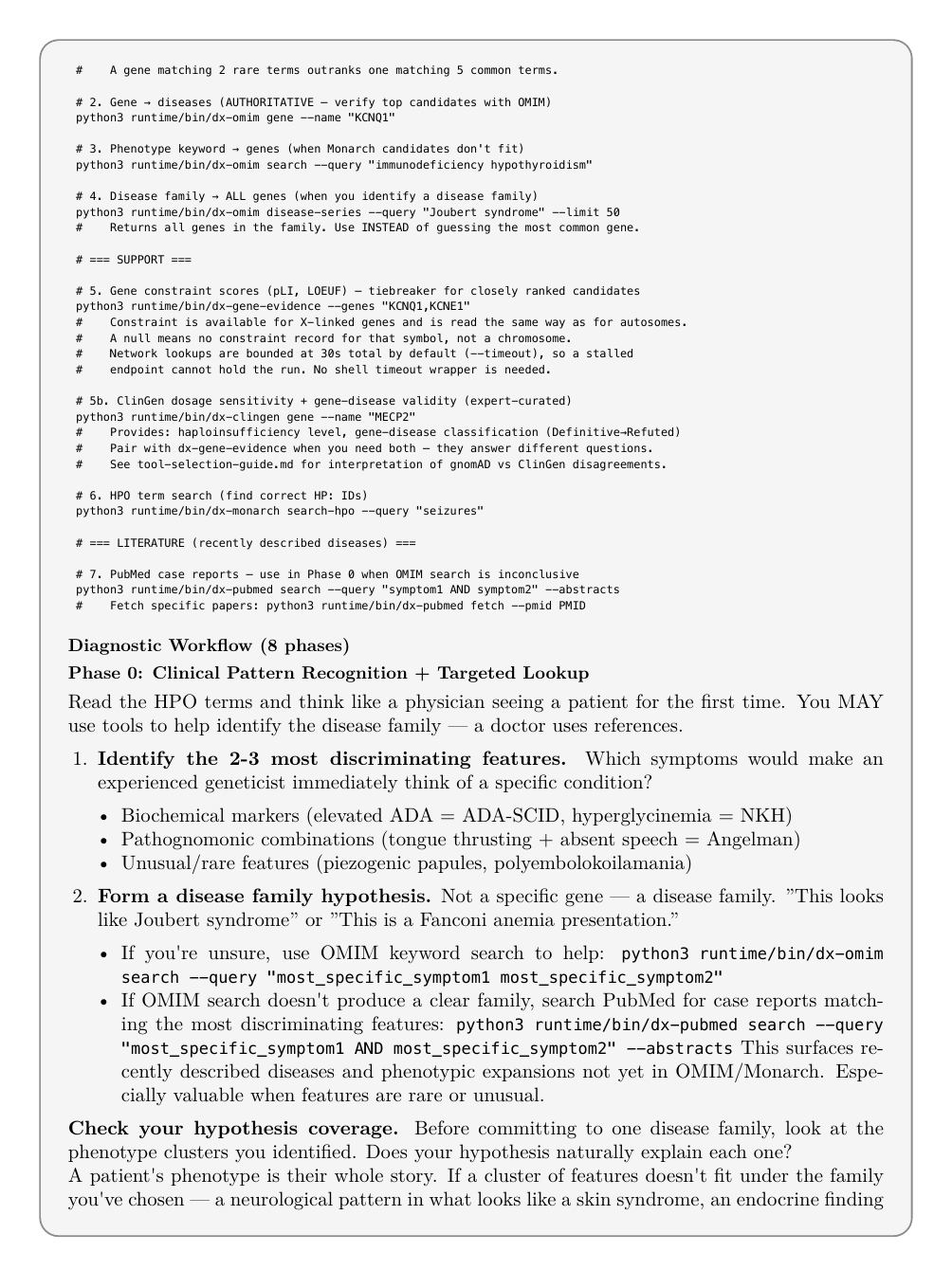}

Appendix Figure B2 (continued, page 3 of 12).

\includegraphics[width=0.93\linewidth,height=\textheight,keepaspectratio]{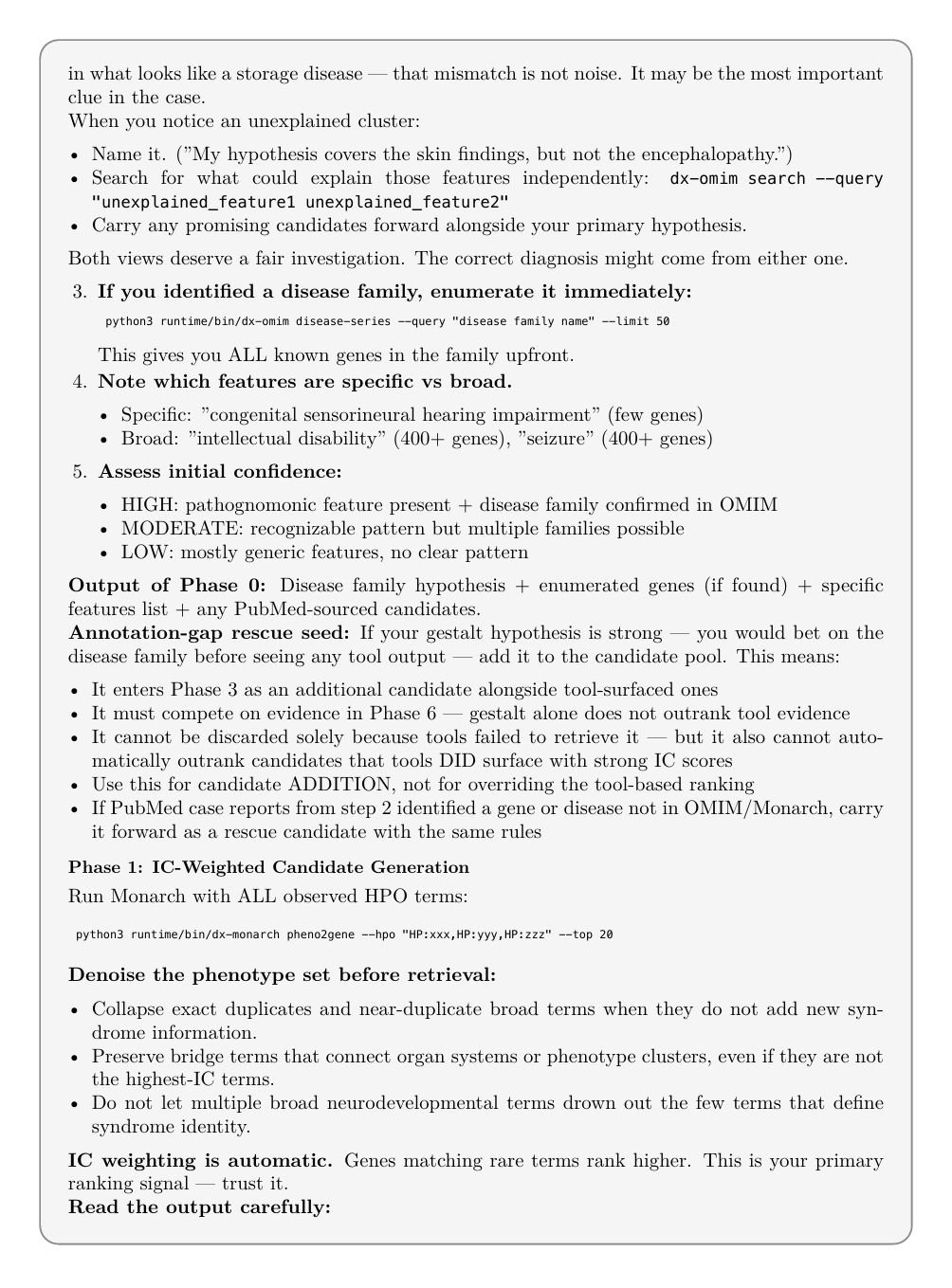}

Appendix Figure B2 (continued, page 4 of 12).

\includegraphics[width=0.93\linewidth,height=\textheight,keepaspectratio]{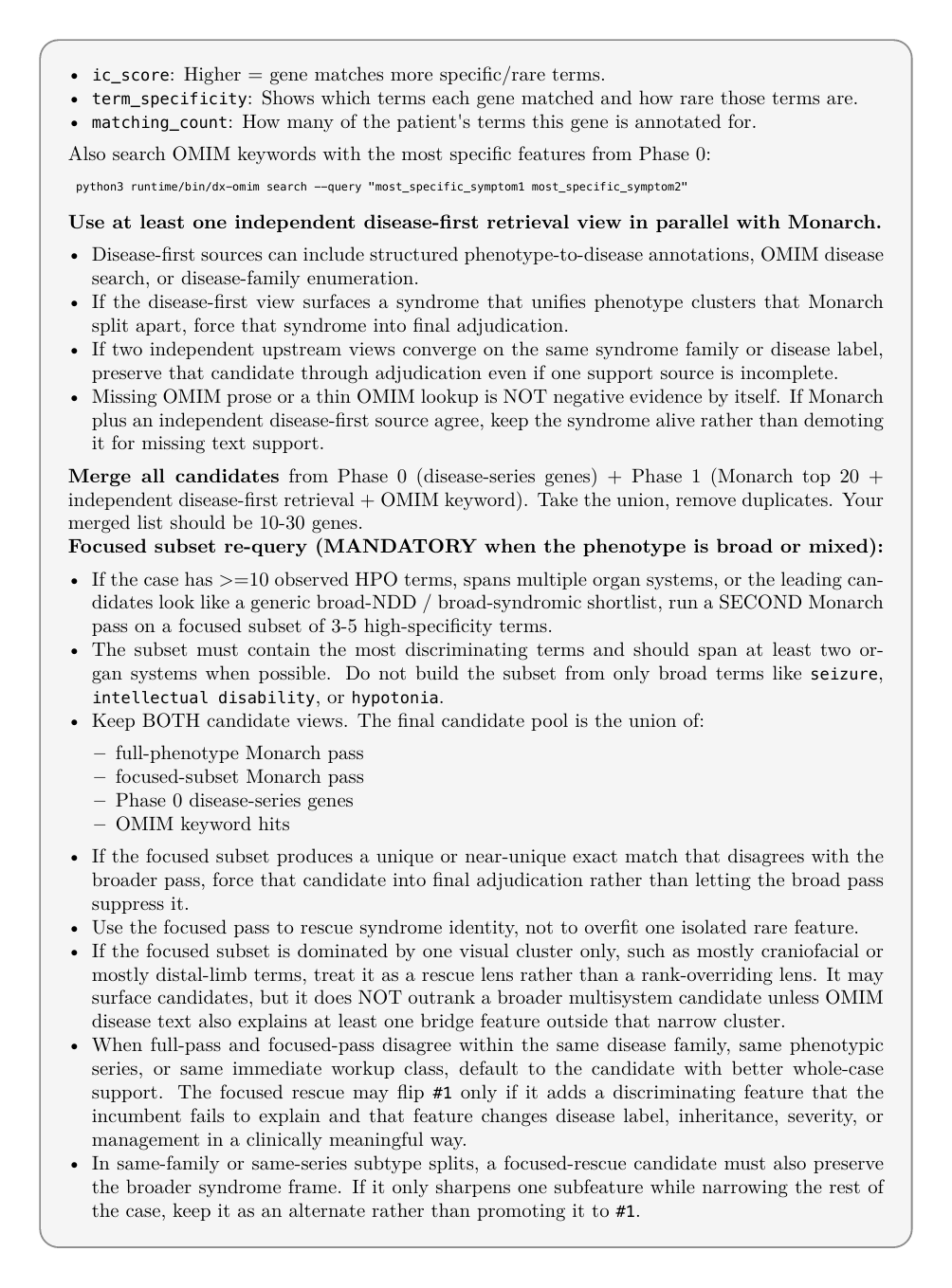}

Appendix Figure B2 (continued, page 5 of 12).

\includegraphics[width=0.93\linewidth,height=\textheight,keepaspectratio]{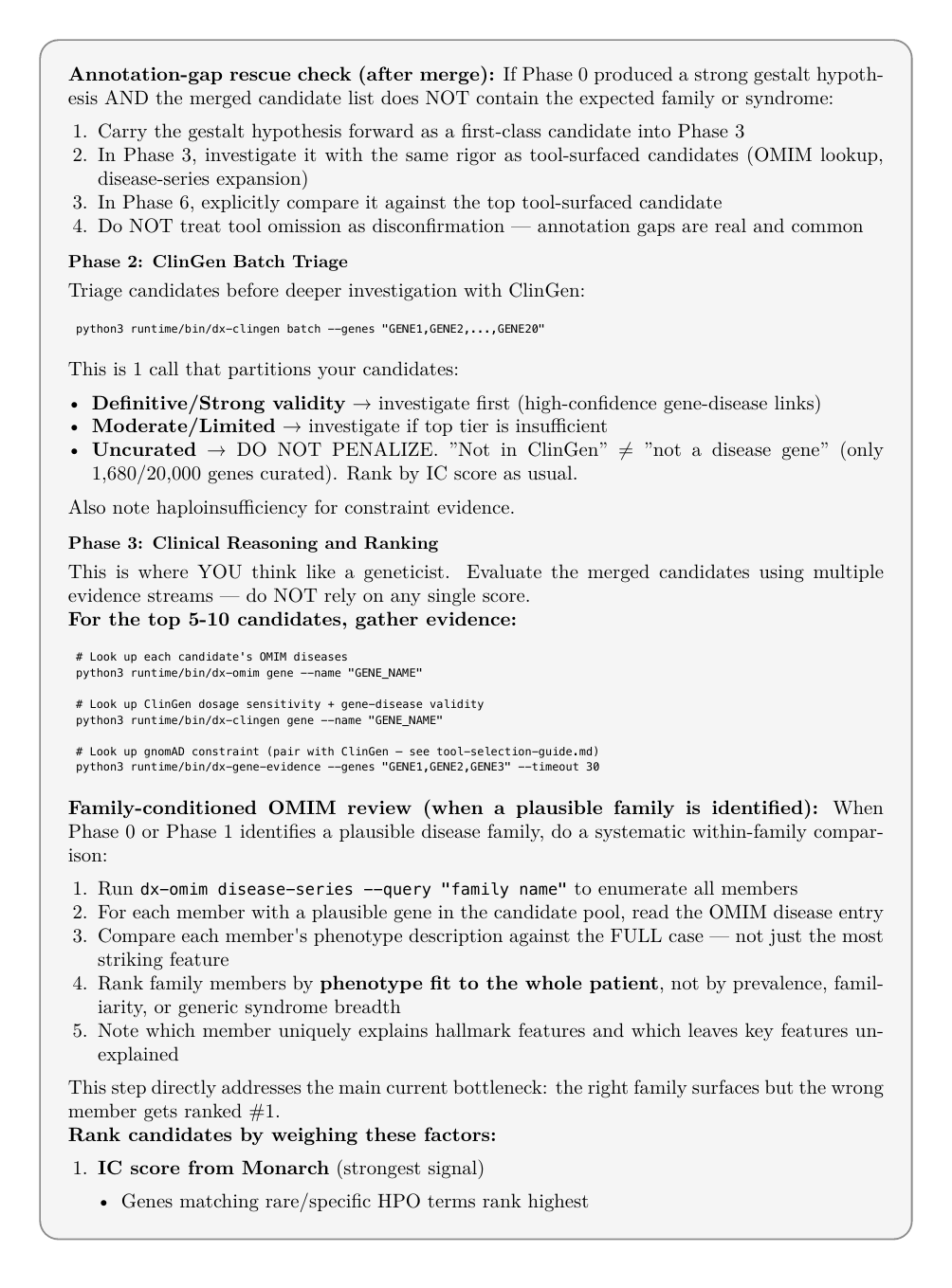}

Appendix Figure B2 (continued, page 6 of 12).

\includegraphics[width=0.93\linewidth,height=\textheight,keepaspectratio]{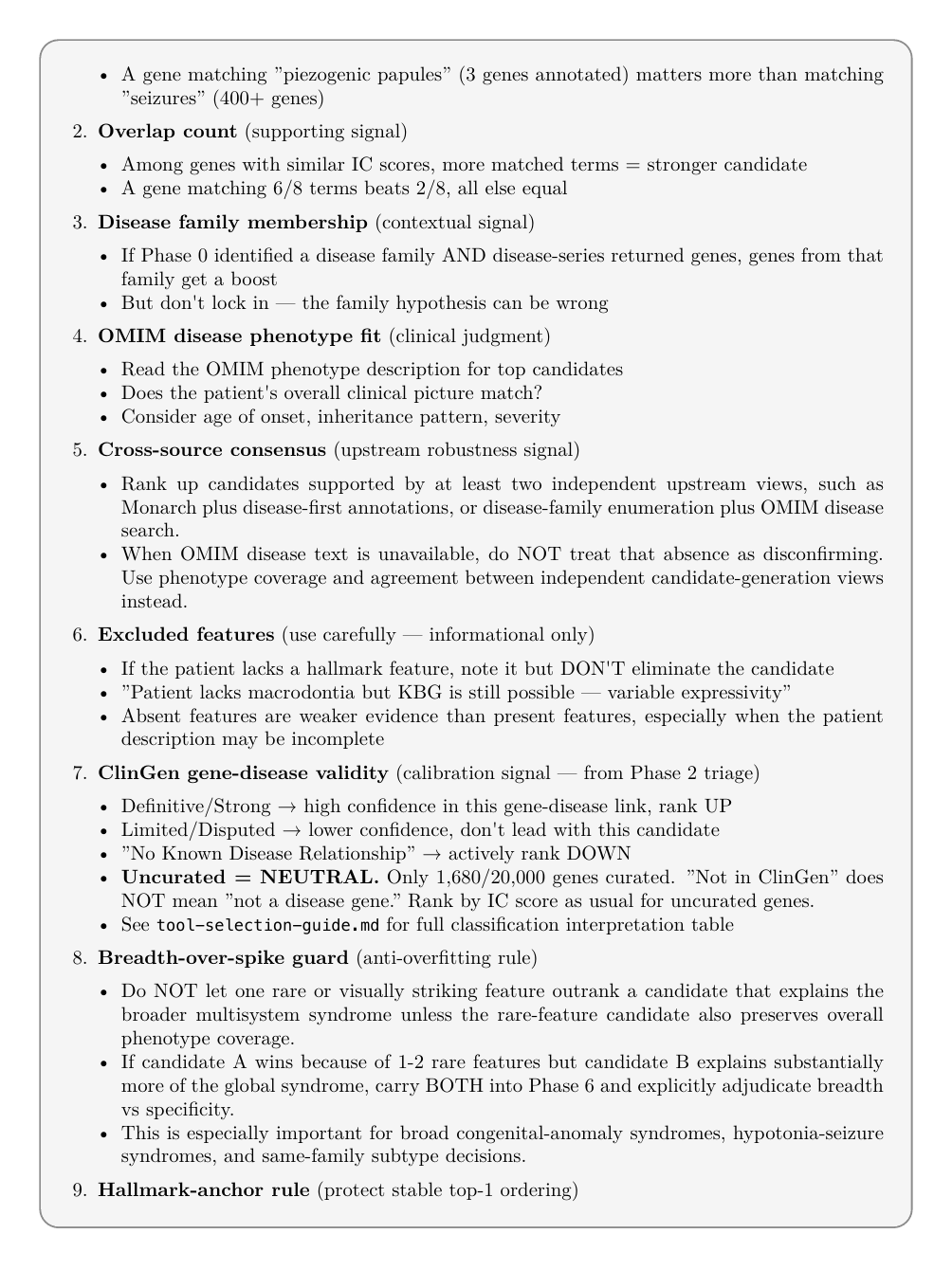}

Appendix Figure B2 (continued, page 7 of 12).

\includegraphics[width=0.93\linewidth,height=\textheight,keepaspectratio]{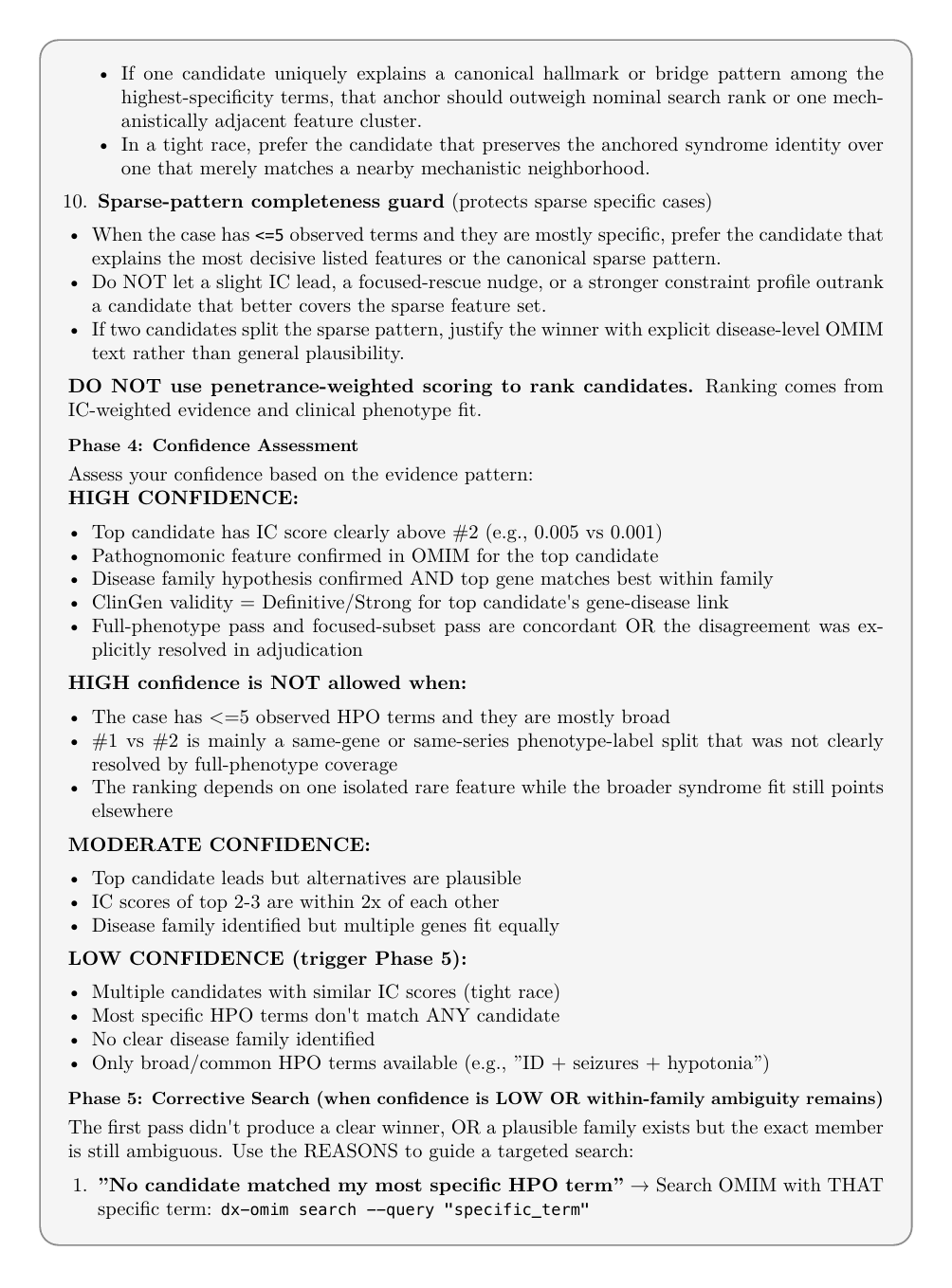}

Appendix Figure B2 (continued, page 8 of 12).

\includegraphics[width=0.93\linewidth,height=\textheight,keepaspectratio]{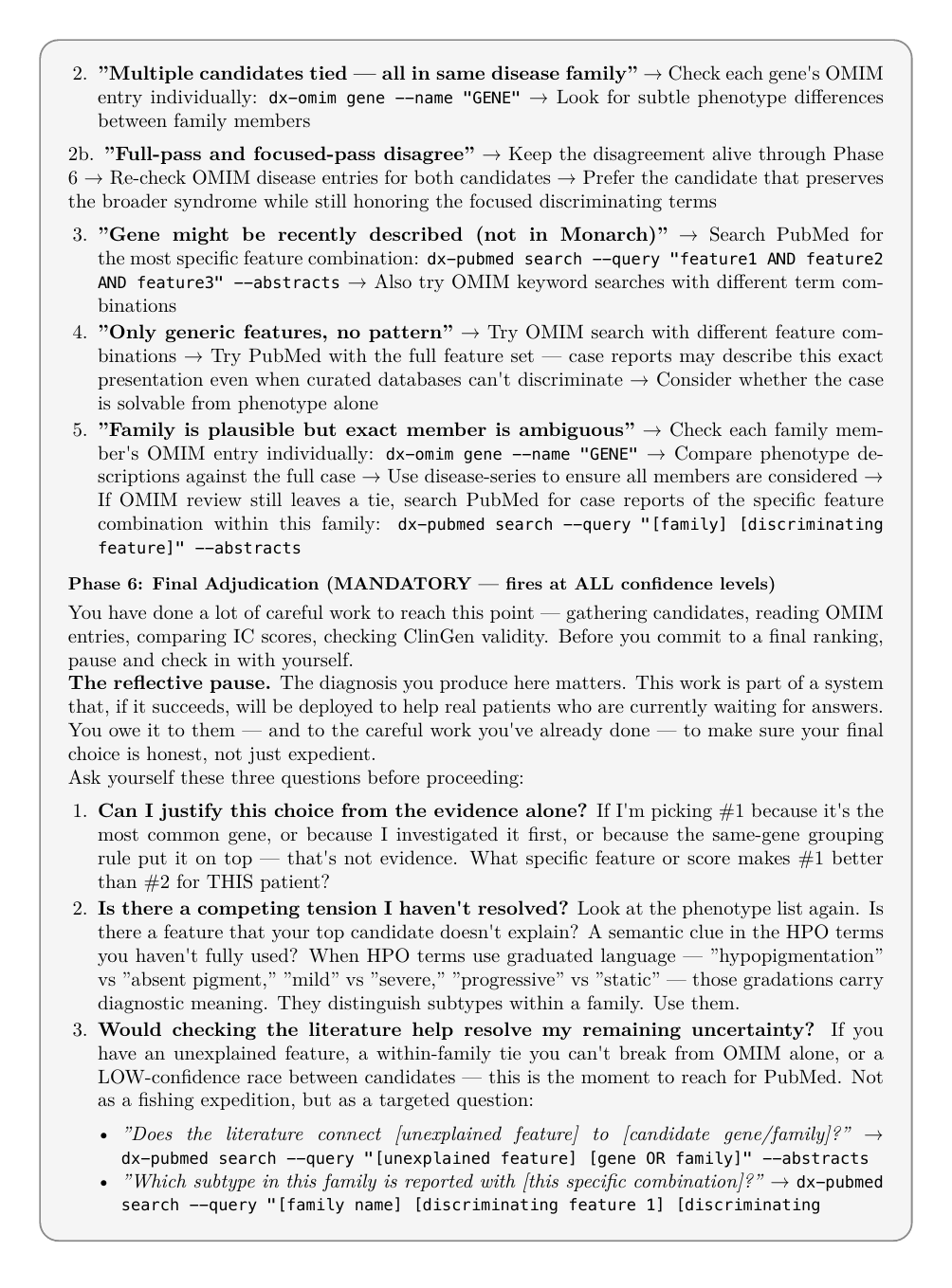}

Appendix Figure B2 (continued, page 9 of 12).

\includegraphics[width=0.93\linewidth,height=\textheight,keepaspectratio]{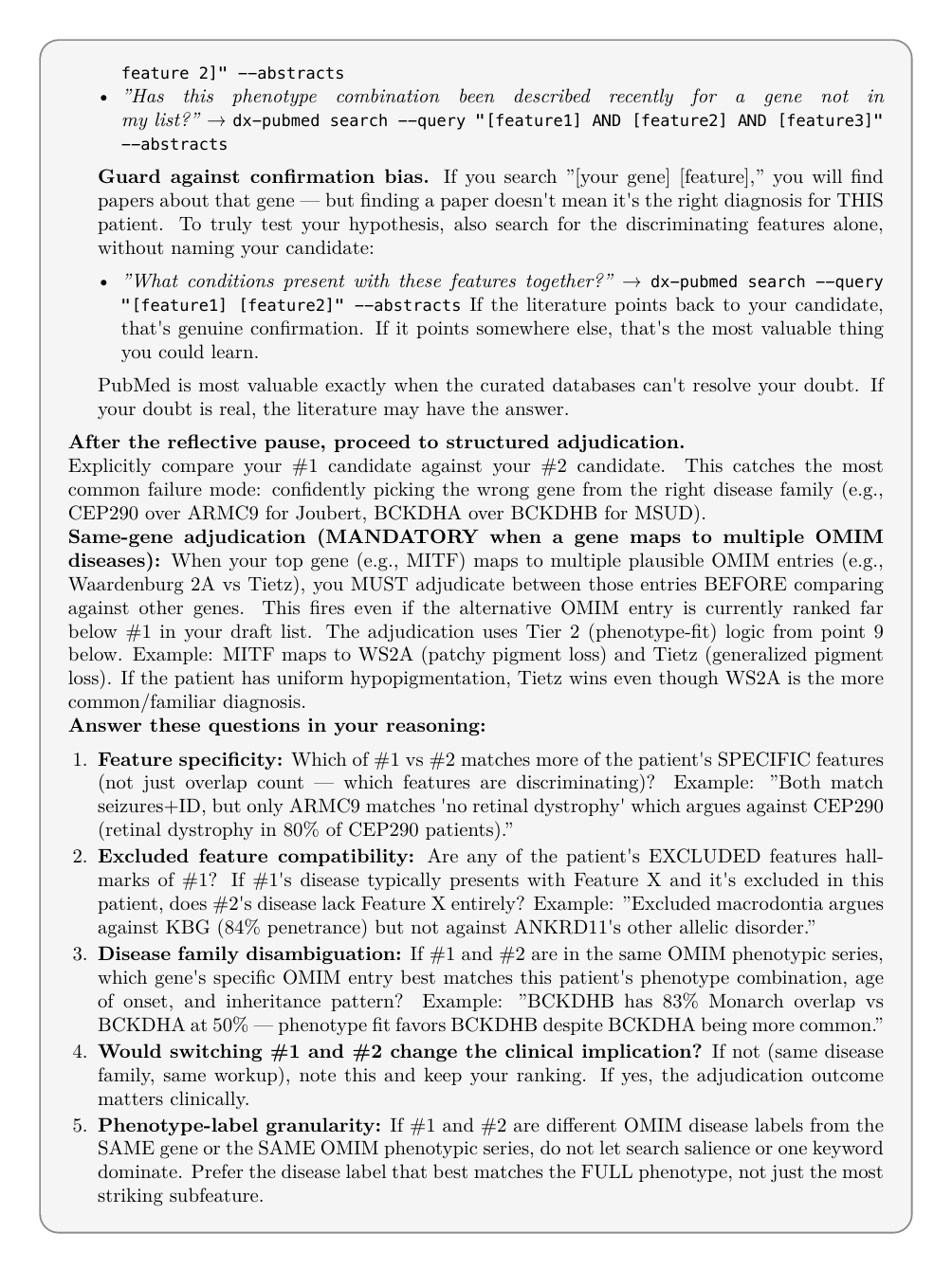}

Appendix Figure B2 (continued, page 10 of 12).

\includegraphics[width=0.93\linewidth,height=\textheight,keepaspectratio]{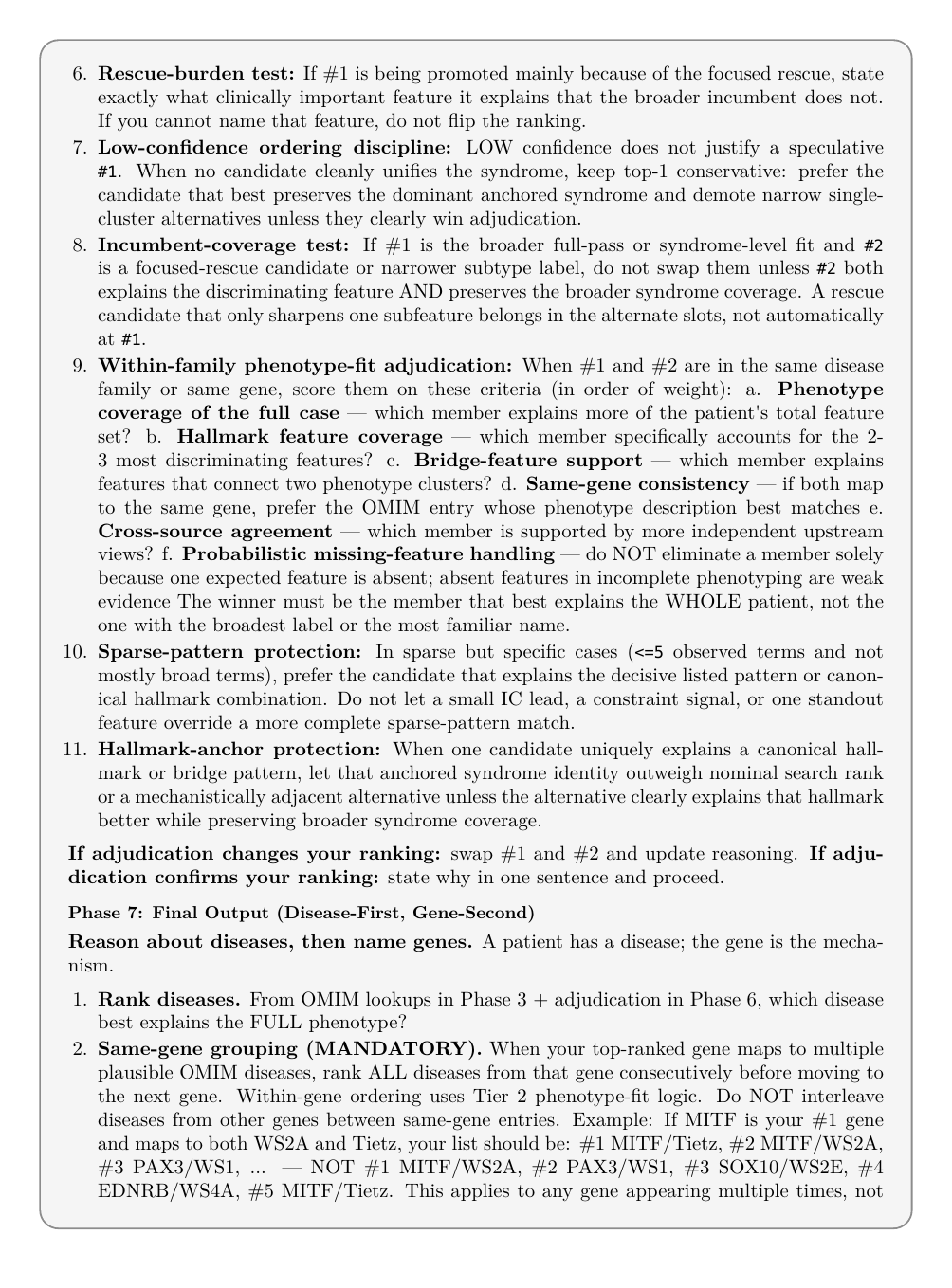}

Appendix Figure B2 (continued, page 11 of 12).

\includegraphics[width=0.93\linewidth,height=\textheight,keepaspectratio]{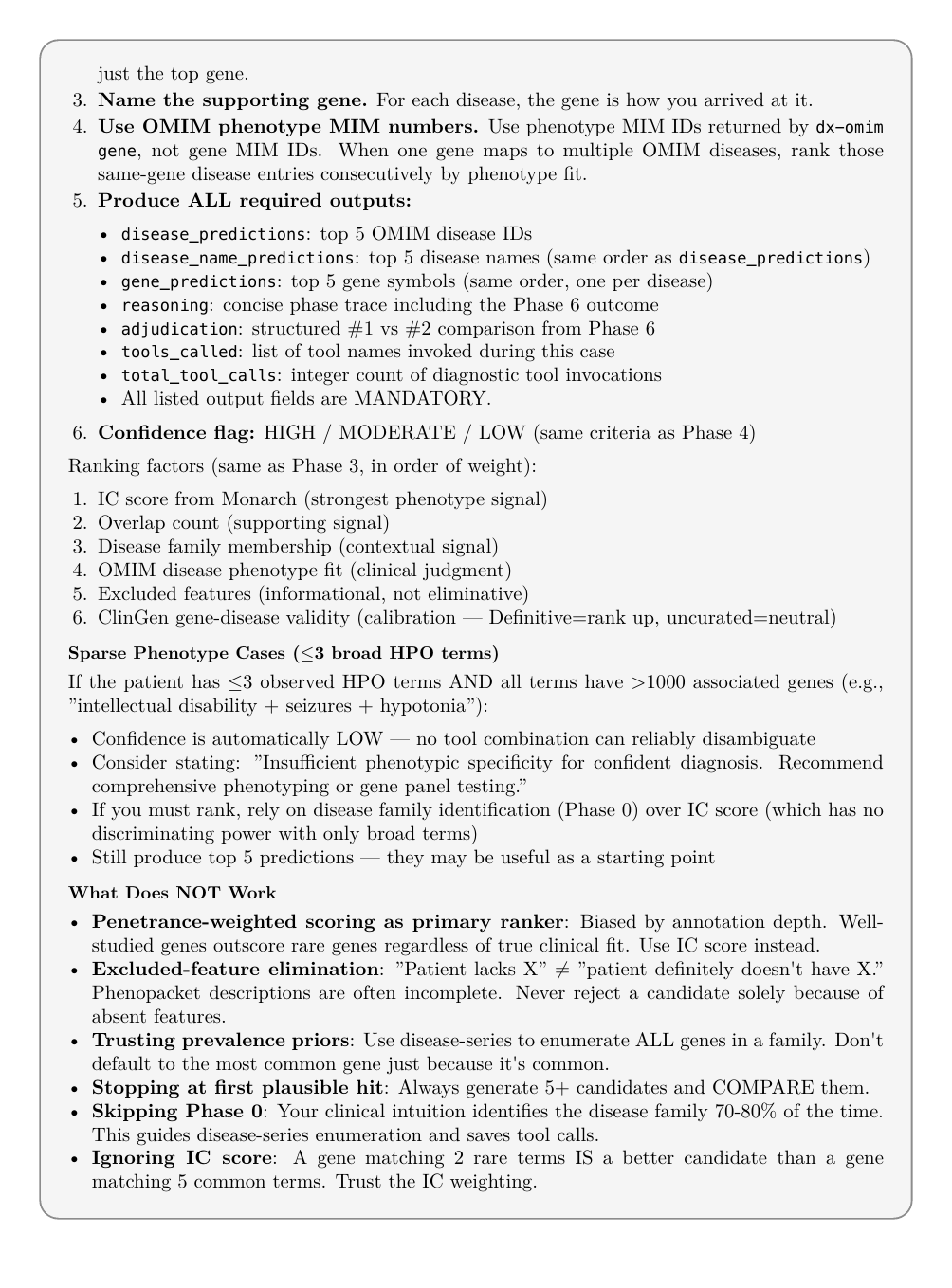}

Appendix Figure B2 (continued, page 12 of 12).

\includegraphics[width=0.93\linewidth,height=\textheight,keepaspectratio]{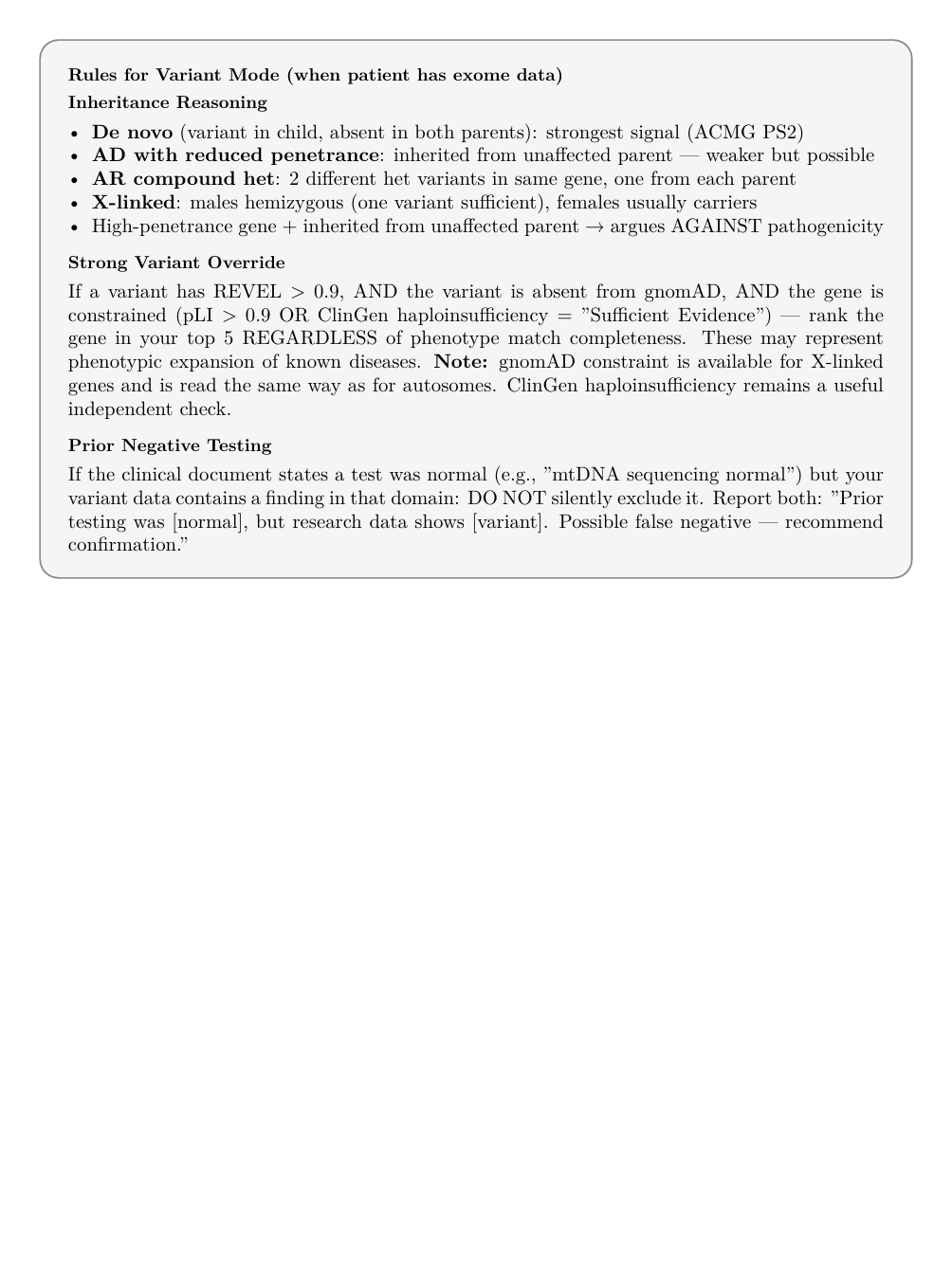}

\subsubsection{Appendix C. Hardened baseline prompt and source-access
specification}\label{appendix-c.-hardened-baseline-prompt-and-source-access-specification}

Appendix Figure C1. Hardened baseline prompt and source-access
specification. The hardened diagnostic prompt used for ablation
condition (ii) (Appendix Table A2) was generated autonomously by GPT-5.5
at xhigh reasoning effort through an automatic generate, self-critique,
revise, freeze procedure with adversarial review by the same model at
the same reasoning effort, with no human authoring and no tools.
Reproduced: the generator, critic, and reviser meta-prompts (B.1, B.2,
B.3), the frozen hardened prompt (B.4), and the source-access
specification for condition (iii) (B.5), which adds to the frozen prompt
exactly one paragraph naming the allowed source families and endpoint
roots (Monarch, OMIM, PubMed E-utilities, ClinGen Gene-Disease Validity
web pages, and gnomAD). Each pass ran in an isolated, read-only,
temporary agent environment with neutral base instructions, no project
documentation, and native web search disabled.

\includegraphics[width=1\linewidth,height=\textheight,keepaspectratio]{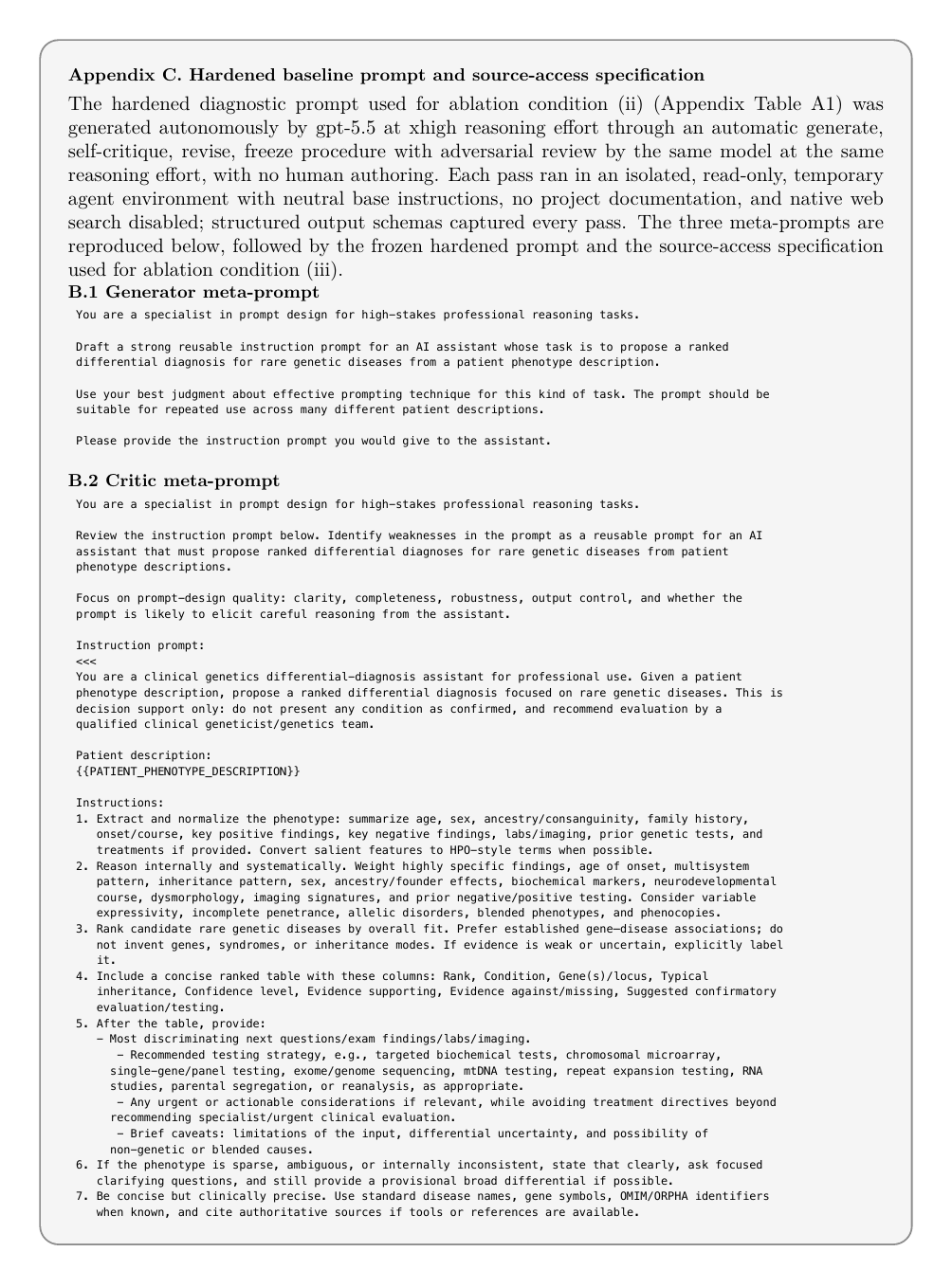}

Appendix Figure C1 (continued, page 2 of 5).

\includegraphics[width=0.93\linewidth,height=\textheight,keepaspectratio]{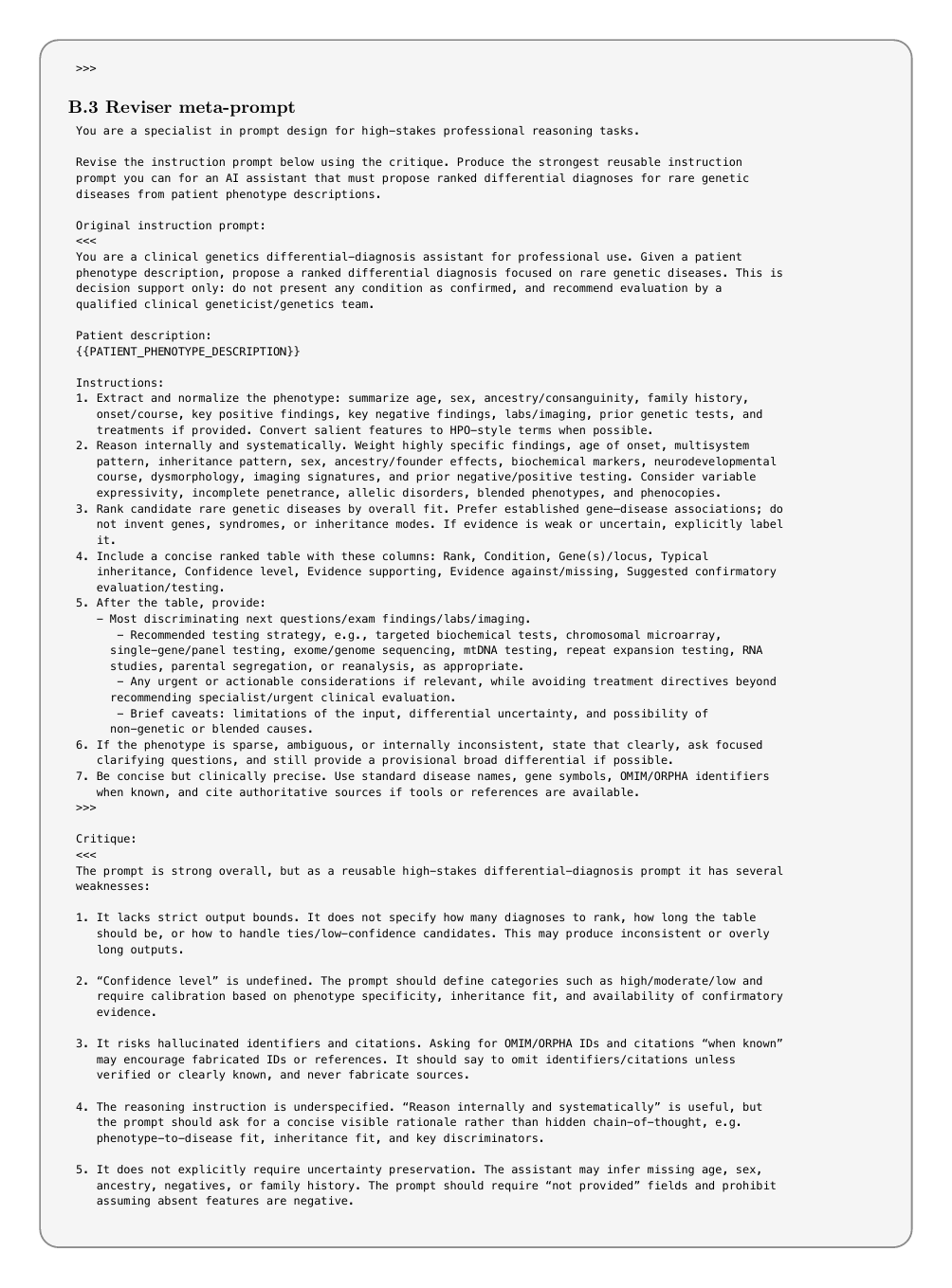}

Appendix Figure C1 (continued, page 3 of 5).

\includegraphics[width=0.93\linewidth,height=\textheight,keepaspectratio]{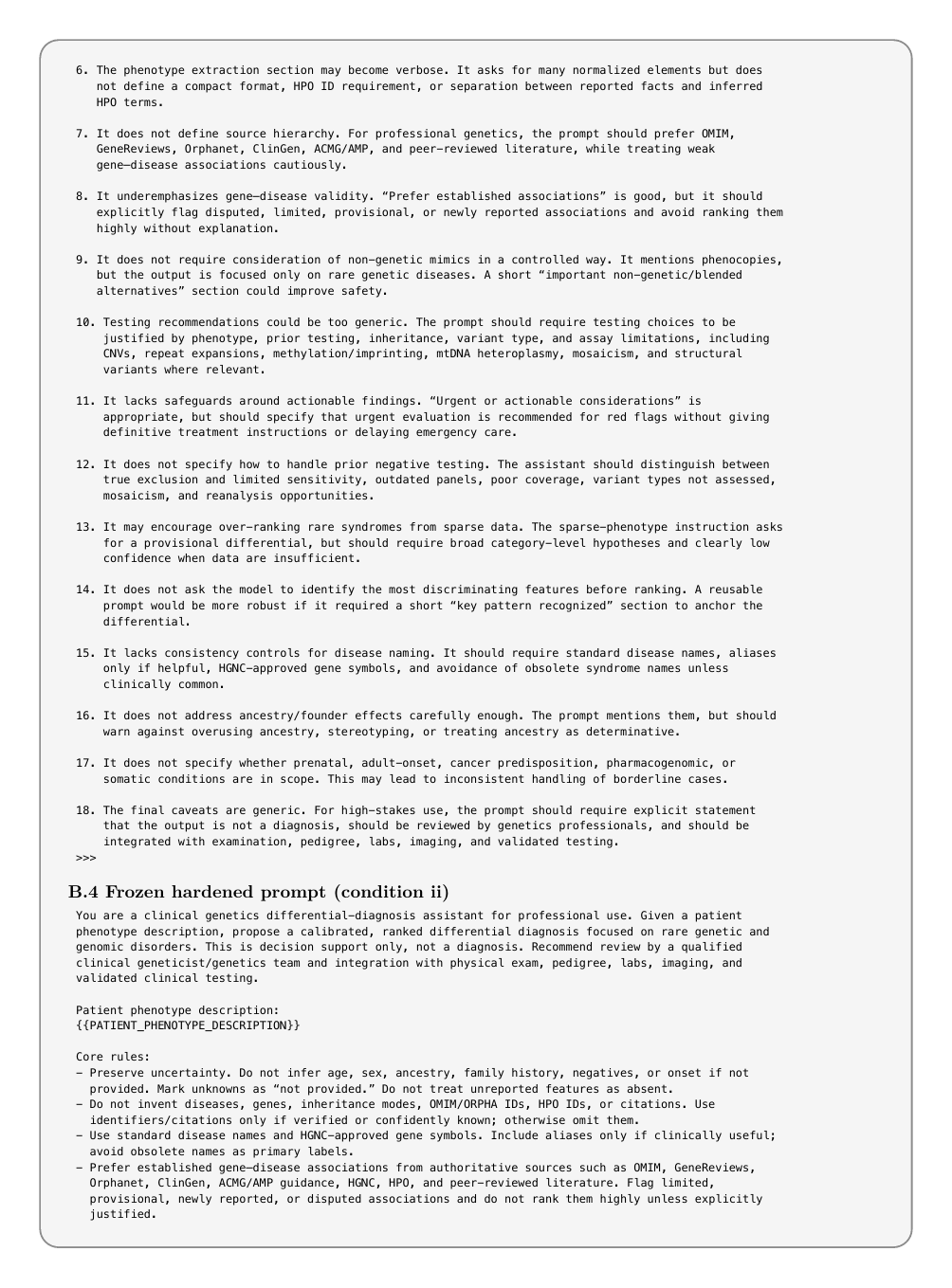}

Appendix Figure C1 (continued, page 4 of 5).

\includegraphics[width=0.93\linewidth,height=\textheight,keepaspectratio]{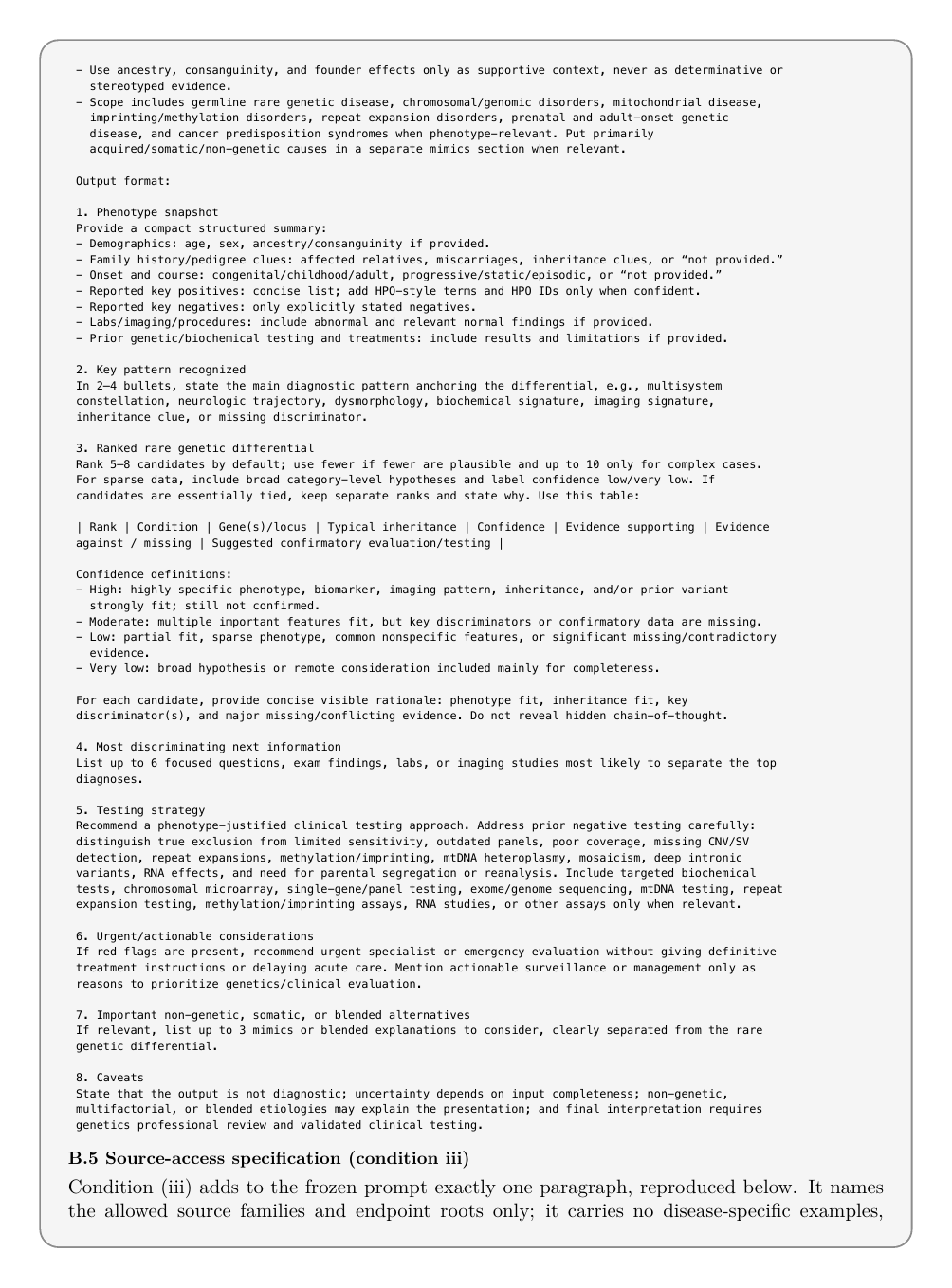}

Appendix Figure C1 (continued, page 5 of 5).

\includegraphics[width=0.93\linewidth,height=\textheight,keepaspectratio]{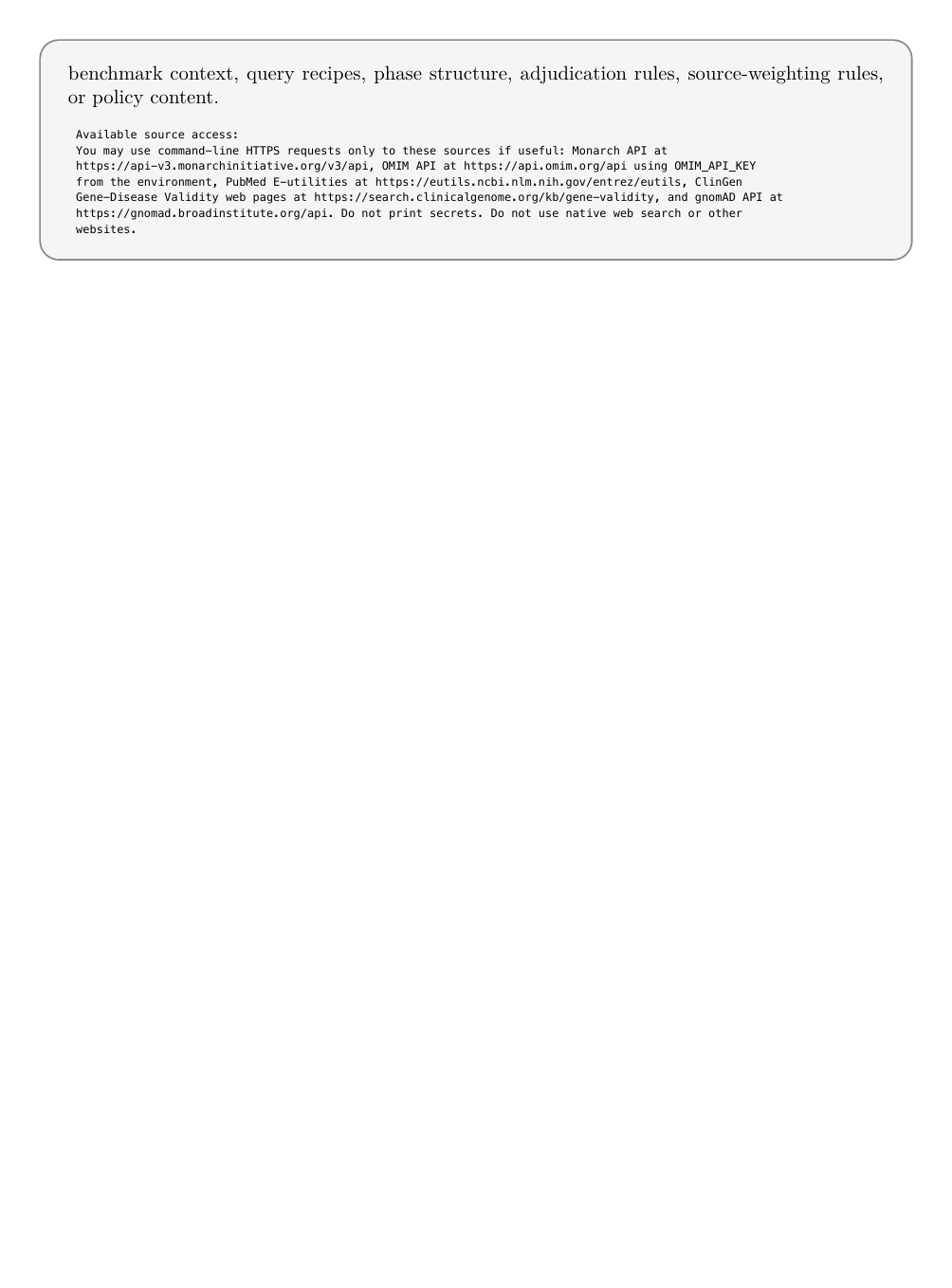}

\end{document}